\documentclass[sn-basic,Numbered,pdflatex]{sn-jnl}
\ifdefined\pdfminorversion\pdfminorversion=7\fi

\usepackage{graphicx}
\usepackage{float}
\usepackage{amsmath,amssymb,amsfonts}
\usepackage[title]{appendix}
\usepackage[table]{xcolor}
\usepackage{booktabs}
\usepackage{array}
\usepackage{xurl}
\usepackage{microtype}
\usepackage{placeins}
\hypersetup{hypertexnames=false}

\newcommand{\cmark}{\ensuremath{\checkmark}}
\newcolumntype{L}[1]{>{\raggedright\arraybackslash}p{#1}}
\newcolumntype{C}[1]{>{\centering\arraybackslash}p{#1}}

\title[SciForge: AI-Native, Multimodal Workbench]{SciForge: An AI-Native, Multimodal Workbench for Scientific Discovery}

\author{SciForge Team, Shanghai Artificial Intelligence Laboratory\\[3pt] \textbf{\small Written by SciForge with DeepSeek-v4-pro (text) and gpt-image-2 (figures), Guided and Verified by Humans}}

\makeatletter
\def\Titlefont{\reset@font\fontsize{17bp}{20bp}\selectfont\raggedright}
\def\Authorfont{\reset@font\fontsize{7.5bp}{9bp}\selectfont\raggedright}
\def\addressfont{\reset@font\fontsize{6.5bp}{7.8bp}\selectfont\raggedright}
\def\abstractfont{\reset@font\fontsize{8.5bp}{10.5bp}\selectfont\leftskip=0pt\rightskip=0pt\parfillskip=0pt plus 1fil}
\def\keywordfont{\reset@font\fontsize{7.5bp}{9bp}\selectfont\leftskip=0pt\rightskip=0pt plus0.5fill}
\long\def\abstract#1{\def\@abstract{\abstractfont #1\par}}
\renewcommand{\@maketitle}{%
  \null\vspace*{-34pt}%
  \hsize\textwidth\parindent0pt%
  \noindent\makebox[\textwidth][r]{\includegraphics[width=0.38\textwidth]{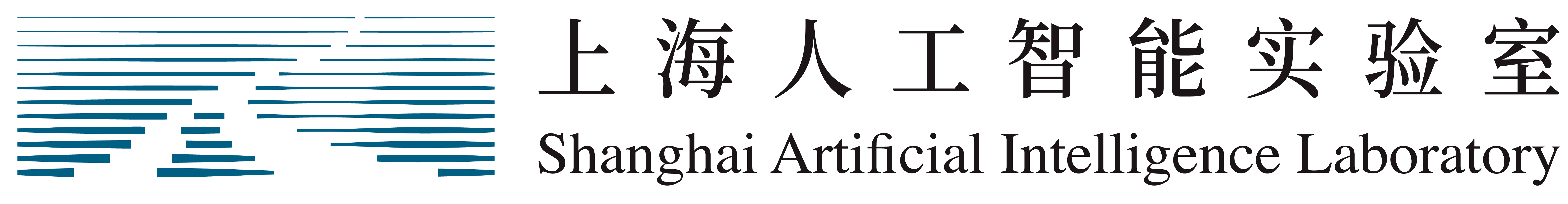}}\par%
  \vspace{4pt}\hrule\vspace{11pt}%
  {\Titlefont\@title\par}%
  \ifnum\aucount>0
    \global\punctcount\aucount%
    \vspace{8pt}{\artauthors\par}%
    \vspace{4pt}{\addressfont\auaddress\par}%
    \ifnum\emailcnt>0\relax%
      \ifx\corrauthemail\@empty\else
        \vspace{6pt}{\addressfont *Corresponding authors. E-mail: \corrauthemail\par}%
      \fi%
    \fi%
  \fi%
  \vspace{10pt}{\printabstract\par}%
  {\printkeywords\par}%
  \vspace{20pt}%
}
\makeatother

\abstract{
Scientific work increasingly spans heterogeneous artifacts---papers, code, datasets, scientific file formats, model outputs, figures, manuscripts, and team decisions---yet general-purpose AI assistants rarely preserve these objects as a coherent, auditable research state. We present SciForge, a multimodal research-native AI workbench that reserves the graphical interface for human judgment while search, parsing, model routing, workflow execution, plotting, writing, and presentation generation run as modular agent-accessible services. SciForge is built around five pillars: (i) \emph{goal-scoped scientific decision governance} for \textbf{goal-oriented} research, with review gates and shared review surfaces; (ii) \emph{translate-then-reason} for \textbf{multimodal} input, routing scientific objects through domain translators before the agent reasons; (iii) \emph{evidence governance} for \textbf{auditable} traceability, linking claims to provenance chains and audit findings; (iv) \emph{collaborative team science} for \textbf{collaborative} research, enabling multi-role decision governance, with shared team workspaces planned for future releases; and (v) \emph{real-world application scenarios} for \textbf{practical} impact, demonstrated through eight end-to-end user cases, with flagship demonstrations including multi-day agentic research sprints for gene discovery, AI-guided de novo protein design, molecular optimization, and genome-to-BGC discovery. The system combines a thin interaction layer, contextual research capability patterns, an Agent Runtime and Workflow Engine, an Evidence-DAG audit sidecar and a Scientific Model Router. SciForge currently runs as a desktop application, with mobile supervision support; future releases will deepen team collaboration. The system is open-source and available at \url{https://github.com/AGI4Sci/SciForge}.
}


\begin{document}

\maketitle

\begin{figure}[H]
\centering
\includegraphics[width=\textwidth]{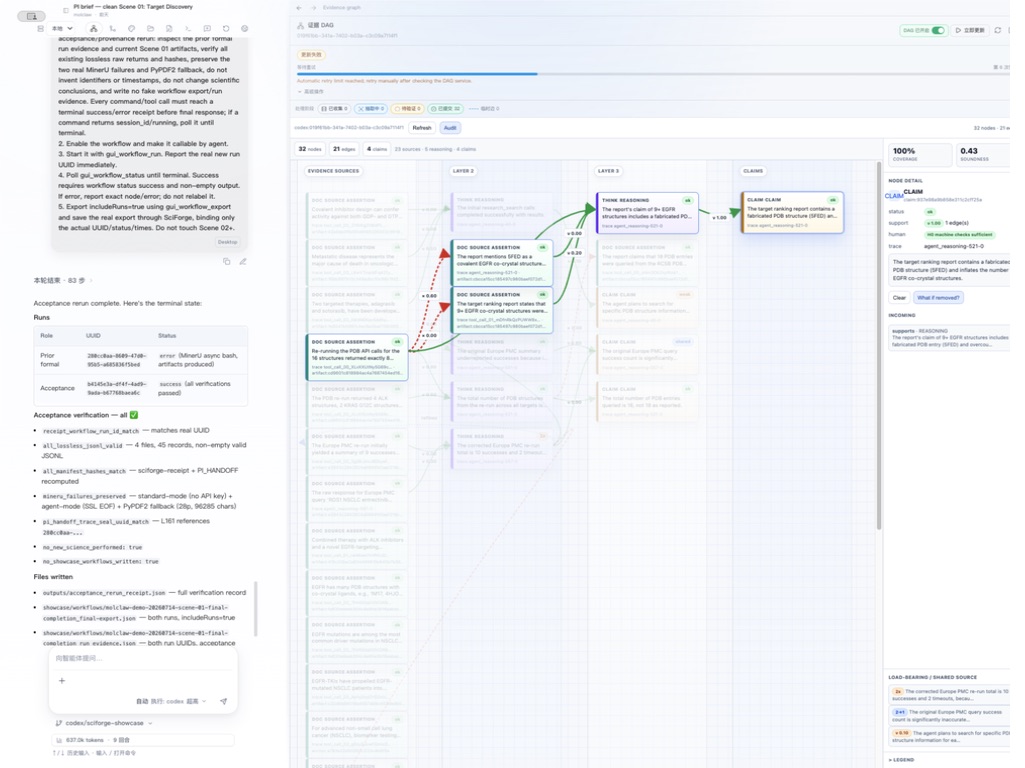}
\caption{SciForge at a glance: a live scientific-research thread (left) and its inspectable Evidence DAG (right). In this target-discovery example, source assertions, reasoning nodes, and claims remain connected to support and contradiction edges, while node-level provenance and audit metrics stay visible to the researcher.}
\label{fig:sciforge-workbench-at-a-glance}
\end{figure}

\vspace{-12pt}
% ============================================================
\section{Introduction}
% ============================================================

\subsection{Background and Motivation}

Scientific discovery is becoming increasingly computational, multimodal, and collaborative. A single research project may move among scholarly papers, experimental protocols, command-line tools, notebooks, gene and variant files, protein sequences, three-dimensional structures, molecular records, single-cell matrices, generated figures, manuscript drafts, slide decks, and group decisions. These objects are not independent files. A manuscript claim may depend on a raw dataset, a preprocessing script, a model run, a visualization parameter, a literature passage, and a later human decision about whether the evidence is strong enough to use. As a result, the central bottleneck is no longer only access to models or tools, but the lack of a persistent environment that can organize scientific objects, agent actions, human decisions, and evidence traces as one coherent research state.

Recent AI systems have demonstrated impressive capabilities for literature search and synthesis, hypothesis generation, coding, experimentation, and scientific writing \cite{paperQA2024,openScholar2024,googleCoScientist2025,sakanaAIScientist2024,yamada2025aiscientistv2,agentLaboratory2025,ghareeb2026robin,huang2025biomni}. At the same time, desktop and terminal-capable scientific workbenches show the importance of keeping execution close to local files, remote servers, and HPC environments \cite{anthropicClaudeScience2026,omicOSOverview2026,omicsClaw2026,operonGitHub2026,claudePrism2026}. However, most existing systems still emphasize either task-specific intelligence, conversational assistance, or local execution surfaces. They rarely treat the full research process as a \textbf{goal-oriented}, \textbf{multimodal}, \textbf{auditable}, and \textbf{collaborative} workflow in which scientific data, agent reasoning, generated artifacts, and manuscript claims remain connected over time.

This gap matters because scientific AI agents must satisfy four interconnected demands absent from general-purpose assistants. \textbf{Goal-oriented} research design requires persistent objectives, review gates, and shared decision records spanning turns, tools, and collaborators---not session-scoped conversations. \textbf{Multimodal} inputs demand native scientific file handling: FASTA sequences, PDB/mmCIF structures, SMILES strings, variant files, and cell-omics tables cannot be directly prompted without domain-specific translation. \textbf{Auditable} evidence traces must link every conclusion to its supporting data, script, model, parameter, paper, or expert translation, and flag when figures or actions require human-PI approval; these requirements align with established FAIR and reproducible-computation principles \cite{wilkinson2016fair,sandve2013reproducible}. \textbf{Collaborative} governance enables team decisions, goal-scoped approvals, cross-session handoffs, and shared review of claims and evidence across multiple researchers. Section~\ref{sec:system-architecture} shows how SciForge's architecture---the Scientific Model Router for multimodal translate-then-reason, the Evidence DAG and Project DAG for auditable provenance, and multi-role decision governance for collaborative science---addresses each demand structurally rather than through ad-hoc prompting.

We present \emph{SciForge}, a multimodal research-native AI workbench for scientific discovery. SciForge is designed as a local-first operating environment for research rather than a wrapper around a single model. Its architecture organizes interaction surfaces, research capability patterns, a core engine with evidence governance, a scientific model router, and local-first infrastructure into a coherent layered workbench; Section~\ref{sec:system-architecture} describes each layer in detail.

\subsection{Contributions}

SciForge's core scientific contributions address four fundamental gaps in current research AI systems:
\begin{itemize}
    \item \textbf{Multimodal Ingress.} End-to-end scientific file processing via translate-then-reason: four classes of native scientific objects---protein sequences (.faa; .fasta/.fa), protein structures (.pdb/.cif/.mmcif), molecules (.smi/.smiles), and single-cell transcriptomics (via Cell2Sentence~\cite{rizvi2025c2s})---are routed through domain-expert translators (Esm2Text~\cite{lin2023esm2}, Prot2Text~\cite{abdine2023prot2text}, BioT5+~\cite{pei2024biot5}, C2S) that return structured expert observations before the main agent reasons. Translator output is an evidence candidate, not a verified fact; the human remains responsible for judgment.

    \item \textbf{Evidence Governance.} Every agent action---file read, model call, tool invocation, script execution---attaches structured provenance (software version, parameters, environment, random seed) to a thread-level Evidence Snapshot. A read-only audit sidecar inspects these snapshots and surfaces potential gaps without blocking the agent's work, linking claims to concrete data, scripts, models, and parameters. A goal-scoped Project DAG organizes research goals, evidence snapshots, and review decisions into a shared, auditable structure supporting candidate and certified release gates, with its provenance vocabulary grounded in the W3C PROV data model \cite{w3cprov2013}.

    \item \textbf{Collaborative Science.} SciForge supports multi-role, collaborative research where reviewer/approver roles, IM-based coordination channels (Zulip, Discord, WeChat, Feishu), and shared review surfaces enable cross-session handoffs and team-level decision governance. The Project DAG makes collaborative review workflows auditable: researchers inspect Evidence Snapshots, record role-attributed ReviewItems, and advance artifacts through candidate/certified release gates with traceable human-PI oversight.

    \item \textbf{Practical Impact.} SciForge is validated through end-to-end scientific workflows spanning biosynthetic gene cluster (BGC) discovery, AI-guided protein and molecular design with experimental evaluation criteria, automated journal-style figure generation, multi-format document parsing, and AI-assisted presentation generation from research artifacts---each with explicit evidence trails, audit findings, and governance checkpoints documented in Section~4.
\end{itemize}

The SciForge codebase, including the Evidence-DAG audit worker, decision-record contracts, agentic skills, and scientific model router, is open source at \url{https://github.com/AGI4Sci/SciForge}.

\section{Related Work}
% ============================================================

\subsection{GUI-Based Scientific Workbenches}

GUI-based scientific workbenches, including desktop and web applications, make AI assistance accessible through human-facing workspaces. Claude Science is the closest broad commercial reference: it is described as an AI workbench for scientists with desktop, local, SSH, and HPC-facing execution, scientific skills, specialist agents, generated artifacts, and reviewer agents for citations and calculations \cite{anthropicClaudeScience2026}. ClaudePrism focuses on the manuscript side of the workflow, combining a native desktop environment with local LaTeX compilation, Python environments, Claude Code sessions, and scientific writing skills \cite{claudePrism2026}. Web tools such as Elicit and Consensus provide polished interfaces for literature search, synthesis, and research-agent style review \cite{elicit2026,consensus2026}. These systems demonstrate the importance of usable research interfaces, but they are usually centered on one model family, one workflow domain, or one user session. SciForge instead treats the GUI as a thin layer for human judgment over a local, modular, multi-agent research substrate.

\subsection{Terminal Workbenches}

Terminal-oriented workbenches prioritize execution close to local files, remote servers, and HPC environments. OmicOS provides a Rust \texttt{omicos-core} binary that can run as a local daemon serving a browser UI or as \texttt{omicos cli}, with analysis code running in a shared IPython kernel and raw data remaining local \cite{omicOSOverview2026,omicOSServe2026,omicOSCLI2026}. OmicsClaw follows a local-first multi-omics design with CLI, TUI, desktop, web backend, and SSH remote execution surfaces over a shared agent loop \cite{omicsClaw2026}. Operon targets bioinformatics workflows across desktop and remote cluster environments, using persistent terminal sessions so jobs execute where data and schedulers live \cite{operonGitHub2026}. These systems are strong at local and remote execution, but they generally expose tools and workflows rather than a unified layer for multimodal scientific routing, research decision surfaces, team supervision, and manuscript-to-artifact continuity.

\subsection{AI Scientists and Research Agents}

Literature-centered agents establish the retrieval and synthesis layer of automated science. FutureHouse exposes agents for literature search, deep review, precedent search, and chemistry planning through a web/API platform \cite{futureHousePlatform2025}, while PaperQA2, OpenScholar, and ScholarQA investigate retrieval-grounded scientific question answering and literature synthesis \cite{paperQA2024,openScholar2024,scholarQA2025}.

End-to-end research agents extend beyond retrieval into ideation, implementation, experimentation, analysis, and manuscript production. The AI Scientist and AI Scientist-v2 automate substantial portions of the machine-learning research loop, Agent Laboratory organizes specialized research roles, and data-to-paper systems turn structured datasets into human-verifiable manuscripts \cite{sakanaAIScientist2024,yamada2025aiscientistv2,agentLaboratory2025,ifargan2025datatopaper}. AI Co-Scientist and Robin study multi-agent hypothesis generation, debate, ranking, and discovery workflows in biomedical settings \cite{googleCoScientist2025,ghareeb2026robin}.

Domain-grounded systems couple language agents to scientific tools or experimental workflows. ChemCrow and Coscientist integrate chemistry planning and execution tools \cite{bran2024chemcrow,boiko2023coscientist}; Biomni, CRISPR-GPT, and the Virtual Lab target biomedical analysis, gene-editing design, and collaborative nanobody design \cite{huang2025biomni,qu2026crisprgpt,swanson2025virtuallab}; SciAgents explores multi-agent graph reasoning for materials discovery, and A-Lab closes part of the loop with autonomous materials synthesis \cite{ghafarollahi2025sciagents,szymanski2023alab}. Benchmarks such as DiscoveryWorld and ScienceAgentBench also show that present agents remain brittle on end-to-end discovery and realistic scientific coding tasks \cite{jansen2024discoveryworld,chen2025scienceagentbench}. Collectively, these systems demonstrate the power of specialized scientific agents, but are less focused on a persistent local workbench in which heterogeneous data, code, figures, decisions, evidence, and generated artifacts remain connected in one governed workspace.

\subsection{Comparison and Positioning}

While each of the three system categories addresses a distinct facet of the AI-assisted research life cycle, SciForge is designed to unify them within a single workbench. The comparison spans five key dimensions: execution, model governance, evidence quality, research continuity, and goal-oriented research design.

\textbf{Execution architecture.} GUI workbenches such as Claude Science and ClaudePrism deliver polished interactive experiences but are typically constrained to one model provider and one local session. Terminal workbenches---OmicOS with its hybrid daemon plus browser model, OmicsClaw with multi-surface CLIs, and Operon with persistent terminal sessions---excel at running code where data and schedulers reside, yet they expose tools and pipelines rather than a unified research workbench. SciForge bridges these worlds by providing a GUI for human judgment while keeping a local-first workspace where execution can span desktop, SSH, and HPC environments.

\textbf{Model governance.} The reviewed public materials commonly emphasize a single provider family or manual switching between separate tools. SciForge's Scientific Model Router combines multi-provider routing with a deliberately narrow translate-then-reason file-ingress path covering four modality classes: protein sequences (Esm2Text~\cite{lin2023esm2}), protein structures (Prot2Text~\cite{abdine2023prot2text}), molecules (BioT5+~\cite{pei2024biot5}), and single-cell transcriptomics (Cell2Sentence~\cite{rizvi2025c2s}); unsupported formats fail closed. Literature is handled through search, PDF anchoring, and source attribution, not as a scientific modality translator. We did not find this same integrated combination documented as a first-class feature in the reviewed categories.

\textbf{Evidence quality and governance.} Benchmarks report substantial failure rates on realistic scientific tasks \cite{jansen2024discoveryworld,chen2025scienceagentbench}; in the publicly documented materials we reviewed, we did not observe the same integrated combination of claim--evidence audit trail, W3C PROV-aligned provenance, and quality gates in a single research workbench. SciForge threads claims through an Evidence DAG backed by a PROV-DM-aligned graph serialized as PROV-JSON \cite{w3cprov2013,huynh2013provjson}, audit runs, risk digests, and explicit quality gates. This governance layer is designed to make agent outputs inspectable and auditable.

\textbf{Research continuity.} Many reviewed systems foreground per-session interaction. SciForge maintains goal-centered research memory as scoped free-text records (scope, tags, confidence, timestamps, context) across sessions; typed goals and decision events belong to the Project DAG, enabling long-running multi-session scientific workflows rather than only single-turn conversations. The combination of research memory, workflow engine, and evidence governance creates a persistent research workspace whose integrated design---tying thread-level Evidence Snapshots to goal-scoped Project Snapshots with audit and release semantics---we did not find documented as a first-class, integrated feature in the public materials of the reviewed GUI, terminal, and API-agent systems.

\textbf{Goal-oriented research design.} SciForge anchors every thread and project to an explicit goal scope with defined review gates, decision records, and release semantics. This design ensures that scientific objectives are persistent, auditable, and governable across multiple turns, tools, and contributors.

Table~\ref{tab:terminal_related_work} summarizes these differences at the category level, while the detailed capability differentiation in Appendix~\ref{tab:capability_differentiation} provides a feature-level comparison against common approaches.

\begin{table}[ht!]
\caption{High-level comparison of scientific AI system categories. A checkmark indicates first-class or documented support as a primary design goal in the publicly reviewed materials. Absence of a checkmark does not prove absence of a feature; it indicates we did not find documented first-class support in the reviewed materials. $\dagger$~Planned for a future release.}
\label{tab:terminal_related_work}
\scriptsize
\centering
\setlength{\tabcolsep}{3.5pt}
\renewcommand{\arraystretch}{1.15}
\begin{tabular}{@{}L{4.6cm} C{1.30cm} C{1.50cm} C{1.30cm} C{1.50cm}@{}}
\toprule
\textbf{Capability} &
\textbf{\shortstack{GUI\\WB}} &
\textbf{\shortstack{Terminal\\WB}} &
\textbf{\shortstack{API\\agent}} &
\textbf{SciForge} \\
\midrule
Human Interface --- GUI for interactive research & \cmark &  &  & \cmark \\
Local Execution --- local/HPC compute, local-first data &  & \cmark &  & \cmark \\
Sci. Tools --- fail-closed native format ingestion &  &  &  & \cmark \\
Sci. Multimodal Routing --- enhanced scientific multimodal understanding &  &  &  & \cmark \\
Workflow Automation --- DAG-based scheduled protocols &  &  &  & \cmark \\
Research Memory --- cross-session knowledge persistence &  &  &  & \cmark \\
Evidence Governance --- provenance, audit trail, release gates &  &  &  & \cmark \\
Team Workspace --- multi-user collaboration\textsuperscript{$\dagger$} &  &  &  & \textsuperscript{$\dagger$} \\
\bottomrule
\end{tabular}
\end{table}

SciForge combines these categories rather than competing with only one of them. It keeps the usability of GUI workbenches, the locality and composability of terminal systems, and the task power of scientific agents, while adding fail-closed scientific-object handling, content-addressed artifacts with structured selection and provenance, specialist domain translation, run-grounded evidence governance with immutable snapshots and an audit sidecar, and goal-scoped decision governance with explicit release semantics.

% ============================================================
\section{System Architecture}\label{sec:system-architecture}
% ============================================================

\subsection{Overall Design Philosophy}

SciForge is organized as a layered research operating environment. \textbf{Layer~1} provides thin judgment surfaces for desktop, group, and mobile interaction. \textbf{Layer~2} encapsulates six research capability patterns---Literature Review, Idea Generation, Experiment Design, Experiment Execution, Analysis \& Iteration, and Scientific Communication---that cover the scientific workflow while sharing one evidence-aware control chain. \textbf{Layer~3} houses the core engine: agent runtime, workflow automation, scoped memory, and evidence governance. \textbf{Layer~4} is the Scientific Model Router that detects modality and dispatches to domain-expert translators. \textbf{Layer~5} supplies infrastructure: local memory, modular workers, scientific connectors, and reproducible run substrate.

SciForge follows three implementation principles that distinguish it from feature-heavy desktop software:

\begin{enumerate}
    \item \textbf{Thin GUI.} The interface appears only when human judgment is needed---reviewing, annotating, approving, comparing; all other capabilities run in the background agent runtime.
    \item \textbf{Modular services.} Search, parsing, plotting, model routing, workflow execution, and presentation generation run as MCP-compatible tools or worker processes, keeping the application close to a thin interaction shell \cite{anthropicMCP2024}.
    \item \textbf{Local-first evidence model.} Data, traces, artifacts, and evidence graphs remain anchored to the research workspace; remote collaboration and mobile access are control channels, not replacements for local state.
\end{enumerate}

\begin{figure*}[ht!]
    \centering
    \includegraphics[width=0.98\textwidth,height=0.86\textheight,keepaspectratio]{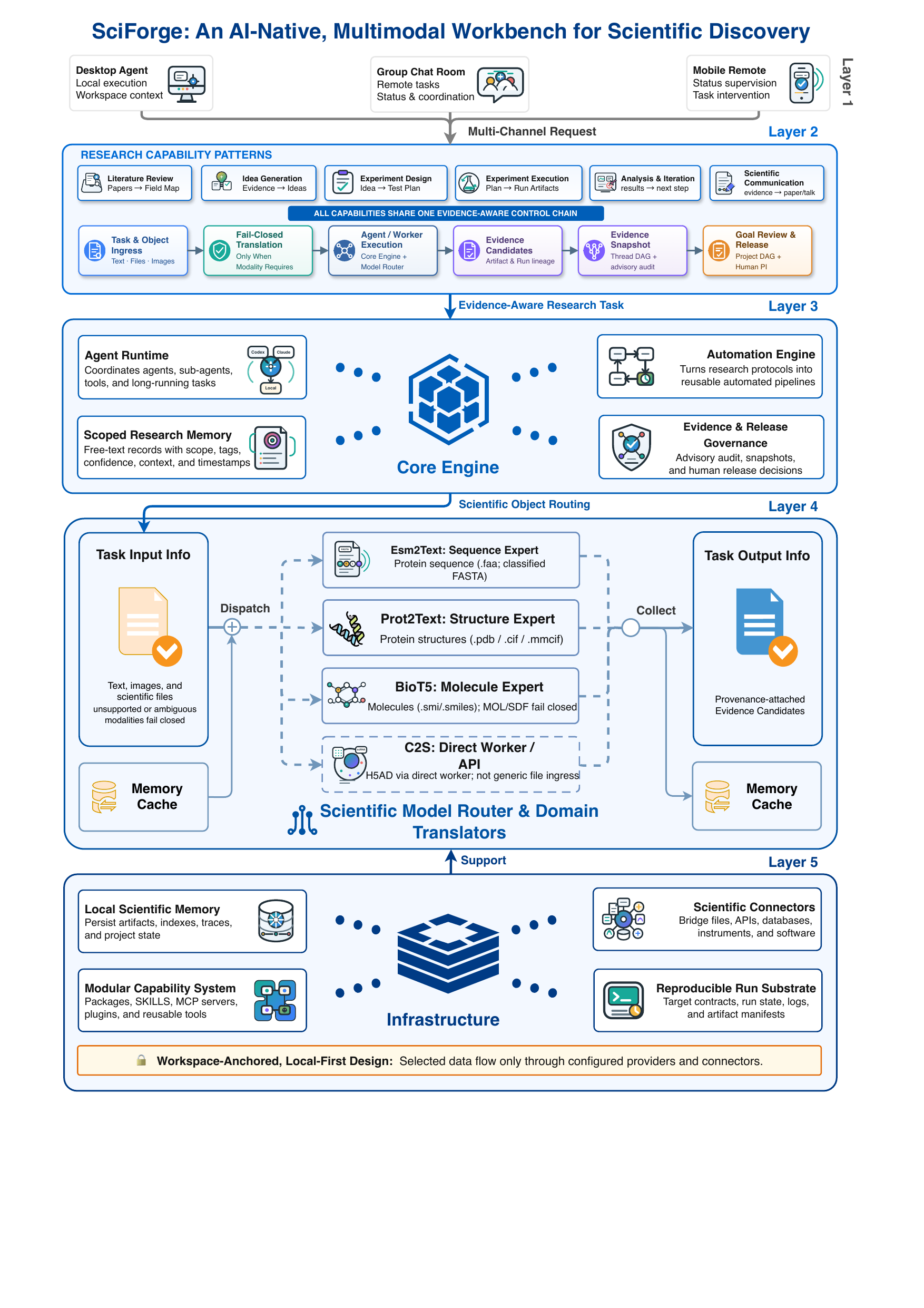}
    \caption{SciForge system framework. Layer~2 exposes six scientific-workflow capabilities---Literature Review, Idea Generation, Experiment Design, Experiment Execution, Analysis \& Iteration, and Scientific Communication. Each capability enters the same evidence-aware control chain---ingress, translation, execution, evidence capture, audit, and release---built on the interaction surfaces, core engine, scientific model router, and local-first infrastructure.}
    \label{fig:sciforge_framework}
\end{figure*}

Figure~\ref{fig:sciforge_framework} summarizes the layered architecture.

\FloatBarrier

\subsection{User Interaction and Collaboration (Layer~1)}
Layer~1 exposes three interaction surfaces driven by the same underlying runtime. The \textbf{desktop application} provides a thin GUI reserved for human judgment---reviewing evidence, annotating figures, comparing candidate artifacts, and approving release decisions---while execution runs in the background. \textbf{Group chat integration} (Zulip, Discord, WeChat, Feishu) enables remote task submission, status polling, and lightweight approvals without switching context. \textbf{Mobile supervision} enables remote monitoring and intervention for long-running experiments, with task submission and review actions gated through the same governance model.

\subsection{Research Capability Patterns (Layer~2)}
SciForge supports six research capability patterns with explicit scientific input--output boundaries. \emph{Literature Review} maps papers and source records to a field map that summarizes prior work and the current state of a domain. \emph{Idea Generation} maps existing evidence to research questions, hypotheses, and candidate methods. \emph{Experiment Design} turns a selected idea into a test plan with validation procedures, metrics, controls, and success criteria. \emph{Experiment Execution} maps that plan to run records and artifacts by executing code, conducting experiments, invoking instruments, or delegating to computational pipelines. \emph{Analysis \& Iteration} maps results to diagnoses, revised assumptions, and a justified next step. \emph{Scientific Communication} maps reviewed evidence and research artifacts to papers, reports, figures, and presentations. These capability patterns cover a common scientific workflow but are not a rigid linear pipeline: researchers may enter at any stage and iterate between stages. They remain contextual patterns within one workbench rather than independent applications with separate execution semantics. Regardless of entry point, each task follows the same evidence-aware control chain: (1) task and object ingress; (2) fail-closed translation when a native scientific object requires a domain expert; (3) delegated agent and worker execution through Layers~3 and~4; (4) provenance-attached evidence candidates with artifact and run lineage; (5) a thread-scoped Evidence Snapshot and advisory audit; and (6) goal-scoped review and candidate/certified release through the Project DAG and human PI.

\subsection{Core Engine (Layer~3)}
The Core Engine plans and executes research work through coordinated subsystems.

SciForge uses two connected evidence layers for scientific governance. The \textbf{Evidence DAG} (thread-scoped) automatically constructs a claim--source--reasoning graph from each agent session's completed turns and persists it as a PROV-DM-aligned graph serialized in PROV-JSON \cite{w3cprov2013,huynh2013provjson}. The \textbf{Project DAG} (goal-scoped) consumes immutable Evidence DAG snapshots across sessions, merges equivalent findings while preserving independent source paths, and links claims to project objectives. Both layers compile asynchronously behind the main agent, keeping evidence up to date without blocking exploration. Audit runs are read-only sidechains that produce structured risk digests---flagging ungrounded claims, unresolved contradictions, weak support, or low-credibility sources---without mutating DAG state. All decisions, whether human or AI, enter through structured Decision Events, giving every modification a traceable, reviewable record.

The \textbf{Agent Runtime} uniformly supports multiple agent backends---Codex, Claude Code, and custom runtimes---allowing research groups to select the execution environment that best matches their scenario: Codex for local-first interactive coding with full workspace access, Claude Code for extended autonomous task chains, and custom runtimes for specialized scientific computing or institutional deployment requirements. Across all backends, the Runtime coordinates the main agent, delegated sub-agents, tool execution, workspace operations, MCP worker invocation, and long-running task recovery through a consistent API surface. The \textbf{Workflow Engine} formalizes reusable pipelines with code, agent, tool, and approval nodes; scheduled execution and loop-style fixed protocol automation are treated as run modes rather than separate architectural modules. \textbf{Scoped Research Memory} stores free-text records (scope, tags, confidence, timestamps, context) across sessions; typed goals and decision events belong to the Project DAG. \textbf{Evidence \& Release Governance} exposes goal-scoped release decisions (candidate and certified gates) through the Project DAG without synchronously blocking exploratory execution.

\subsection{Scientific Model Router (Layer~4)}
The \textbf{Scientific Model Router} is a core architectural innovation of SciForge, acting as a universal model invocation plane that abstracts away provider heterogeneity while retaining the full expressiveness of each backend. Rather than forcing a lowest-common-denominator interface, the Router supports three request--response protocols---OpenAI Chat Completions, OpenAI Responses, and Anthropic Messages---selected per provider through a normalized endpoint format, allowing each model to be invoked through its native, most capable API surface.

\textbf{Intelligent Auto-Routing.} The Router includes a lightweight classifier that inspects the user request and recent context to automatically select the appropriate reasoning effort (off, high, or max), with a deterministic heuristic fallback so routing decisions do not rely on a single point of failure.

\textbf{Provider-Aware Optimization.} For DeepSeek models, the Router performs proactive reachability probing, applies native thinking controls, and computes real-time cache-aware cost estimation in both USD and CNY. For all providers, it implements automatic retry with backoff, error classification, and stream-idle timeout detection.

\textbf{Output Robustness and Repair.} The Router repairs malformed tool-call arguments---stripping Markdown fences and recovering partial JSON---and extracts embedded images from tool results as standard model attachments, enabling multimodal reasoning chains across scientific visualizations and model analysis.

\textbf{Scientific Modality Routing.} For scientific files, the Router detects modality-specific extensions and applies one of two handling paths. \emph{Translate-then-reason} applies to four currently supported modalities: protein (.faa; .fasta/.fa with per-record conservative content classification) translated by Esm2Text, protein structures (.pdb, .cif, .mmcif) by Prot2Text, small molecules (.smi, .smiles) by BioT5+, and single-cell transcriptomics via Cell2Sentence (C2S). Each translator output carries provenance metadata, allowing the main agent to reason over structured expert observations. VCF, BED, GFF, and MGF are unsupported and fail closed; raw content is not routed to the general reasoner.

Multimodal translation caches store scientific and visual observations by content hash, enabling repeated requests to reuse expert evidence without recomputing final conclusions.

\subsection{Infrastructure (Layer~5)}
The Infrastructure layer provides the durable substrate that anchors research activity to the local workspace. \textbf{Local Scientific Memory} stores papers, datasets, parsed documents, scientific file manifests, embeddings, indexes, tool outputs, run metadata, and artifact caches. The \textbf{Modular Capability System} packages computational functionality as Skills, MCP servers, Plugins, HTTP services, or worker processes---each independently versioned and upgradeable. \textbf{Scientific Connectors} normalize external resources such as papers, datasets, databases, APIs, instruments, notebooks, analysis scripts, reference managers, LIMS/ELN systems, and institutional storage into a unified access layer. The \textbf{Reproducible Run Substrate} records execution-target contracts, run and job state, logs, and artifact manifests so that downstream evidence can refer to concrete execution records. SciForge remains workspace-anchored and local-first; selected data may flow through explicitly configured model providers, connectors, or remote execution services.

% ============================================================
\section{Use Cases}
% ============================================================

\subsection{Agentic Research Sprint: From Question to Manuscript Package}

\begin{figure*}[ht!]
    \centering
    \includegraphics[width=0.96\textwidth,height=0.35\textheight,keepaspectratio]{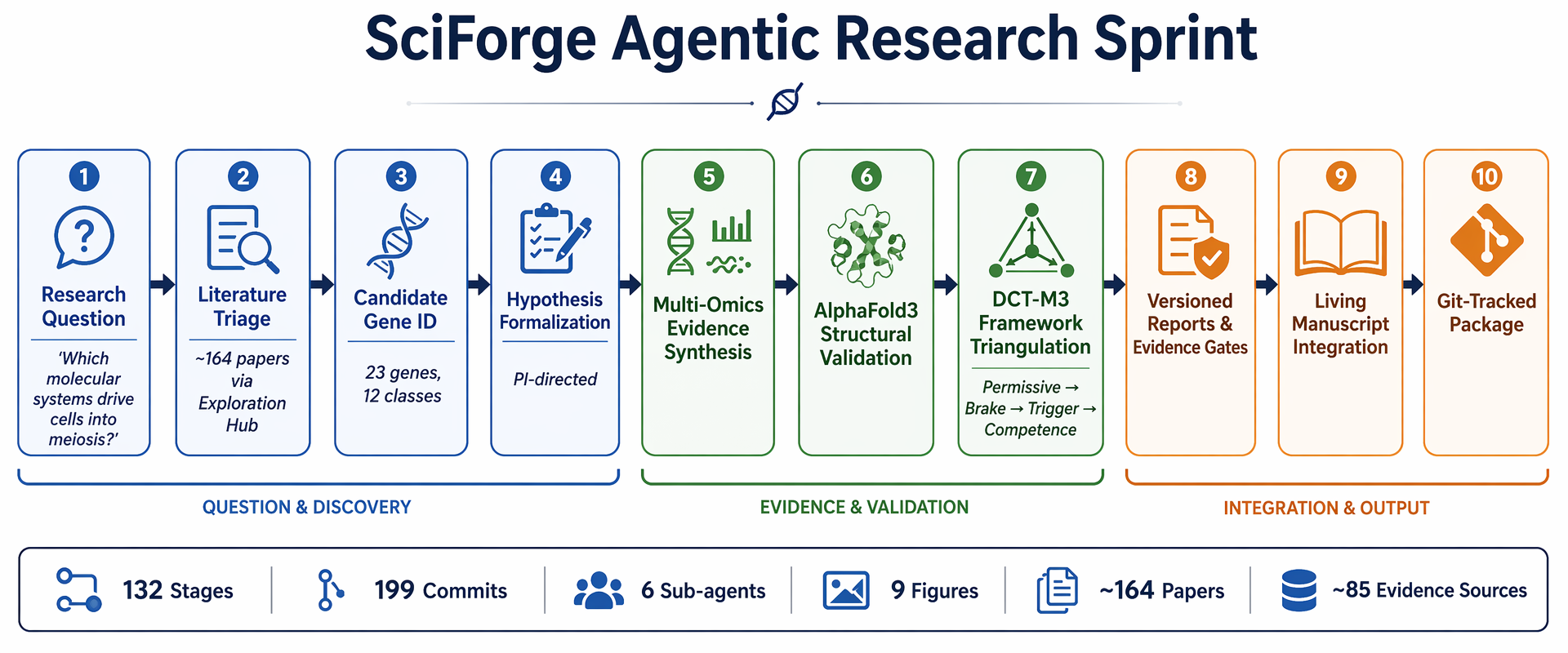}
    \caption{Automated research pipeline executed during the SciForge scenario-01 research sprint. The 132-stage, 199+-commit agentic loop progresses from a biological research question through literature mining (DCT-M3), candidate gene curation, AlphaFold3 structural prediction, evidence triangulation, manuscript synthesis, and submission packaging. Quantitative metrics: 132 research stages, 199+ Git commits, 6 parallel child-agent runs, 9 generated figures, $\sim$164 papers reviewed, and $\sim$85 evidence nodes produced. Selected artifacts and thread evidence are linked to versioned records, evidence summaries, and audit findings where captured.}
    \label{fig:research-pipeline}
\end{figure*}

\textbf{Setting.} A human PI poses the biological question ``Which molecular systems drive mammalian germ cells into meiosis?'' and delegates execution to SciForge, an AI4S research workbench. Codex---an AI coding agent replacing the human expert role---orchestrates SciForge's agent-accessible services---literature search, candidate gene identification, structured hypothesis formulation, parallel sub-agents for multi-omics evidence synthesis, AlphaFold~3 structural assessment \cite{abramson2024alphafold3}, and iterative methodology hardening---within a multi-day, PI-controlled agentic loop (Fig.~\ref{fig:research-pipeline}), with Codex serving as the PI's AI delegate. Outputs are reviewed through evidence summaries, reviewer-style self-audits, and human decisions. After 132 stages and 199+ Git-tracked commits, the repository\footnote{\url{https://github.com/AGI4Sci/scenario-01-research-sprint} (archive available at time of writing)} records the control loop as a local workspace log.

\begin{figure*}[ht!]
    \centering
    \includegraphics[width=0.85\textwidth,keepaspectratio]{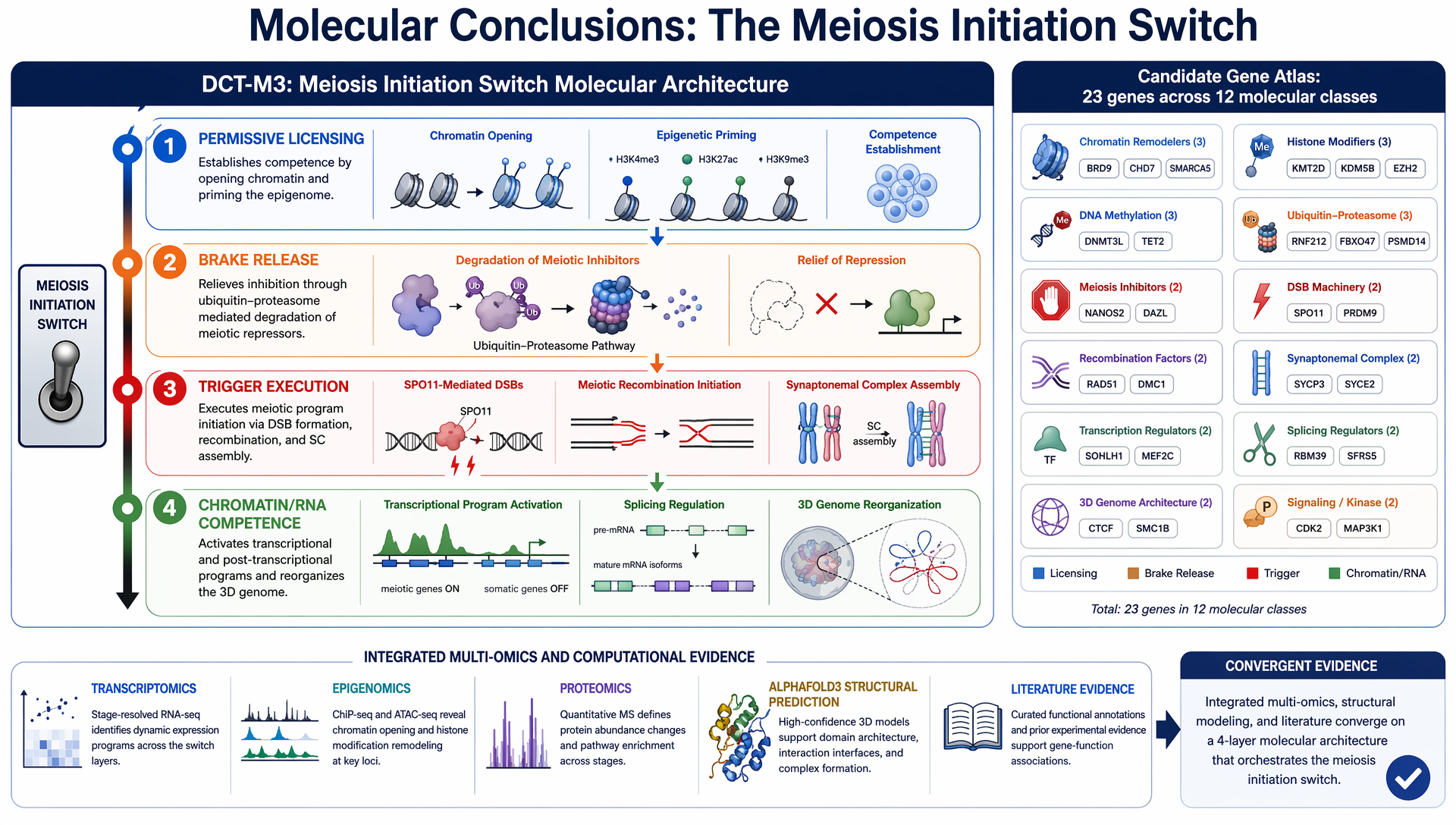}
    \caption{Key biological conclusions reported by the SciForge agentic research sprint. The DCT-M3 causal-module triangulation framework decomposes the meiosis initiation switch into four regulatory layers: (1)~permissive licensing---chromatin opening and epigenetic priming; (2)~brake release---degradation of meiotic inhibitors via ubiquitin-proteasome pathways; (3)~trigger execution---SPO11-mediated double-strand breaks and synaptonemal complex assembly; (4)~chromatin/RNA competence---transcriptional activation, splicing regulation, and 3D genome reorganization. The sprint reported a candidate-gene atlas of 23 genes across 12~molecular classes, supported by multi-omics evidence, AlphaFold3 structure-based computational assessment, and literature triangulation; formal expert adjudication remains future work.}
    \label{fig:meiosis-conclusions}
\end{figure*}

\textbf{Discovery.} The systematic agentic pipeline yielded a candidate-gene atlas of 23 genes across 12 molecular classes, with claims traceable to specific evidence sources (Fig.~\ref{fig:meiosis-conclusions}). Central to the discovery was the DCT-M3 causal-module triangulation framework, which decomposes the meiosis initiation switch into four regulatory layers: (1)~permissive licensing---chromatin opening and epigenetic priming via pioneer transcription factors; (2)~brake release---degradation of meiotic inhibitors through ubiquitin-proteasome pathways; (3)~trigger execution---SPO11-mediated programmed double-strand breaks and synaptonemal complex assembly; and (4)~chromatin/RNA competence---transcriptional activation, alternative splicing, and 3D genome reorganization. Multi-omics evidence (transcriptomics, epigenomics, proteomics) was cross-validated against AlphaFold3 structure-based computational assessments and systematic literature triangulation to ground each candidate in causal evidence. This scenario demonstrates that SciForge can sustain long-horizon, evidence-constrained research---not by auto-generating a paper, but by producing an inspectable, auditable research package whose reasoning and evidence trail remain open to human scrutiny.

\textbf{Informal Post-Hoc Literature Mapping.} A post-hoc review of the 23-gene atlas against the published meiosis literature found that 22 of the 23 genes have known meiosis associations; MAPK fell outside established meiosis paradigms. This retrospective mapping was conducted informally and was not guided by a pre-registered protocol. Formal expert adjudication of gene-level relevance, along with inter-annotator agreement metrics and a pre-registered evaluation protocol, remains future work.

\FloatBarrier
\subsection{AI4AI: Automated Modeling for Scientific Discovery}

AI-for-AI (AI4AI) refers to a design paradigm in which agents read, distill, and recombine algorithmic building blocks from existing codebases into new machine learning models. Rather than treating model design as a one-shot prompt, AI4AI treats it as an iterative design loop: an agent reads external repositories, extracts reusable algorithm atoms (e.g., attention mechanisms, distribution losses, latent regularizers), maps them to the target project's training interface, composes a new model, and submits training, inference, and evaluation jobs through a computing-platform job-submission system. Evaluation metrics then feed back into the next design cycle.

We exercised this paradigm on protein contact prediction---the task of predicting residue--residue contacts from amino acid sequences using a pre-trained protein language model. The agent was given a lightweight codebase centered on a \texttt{ContactProbe}---a small linear or MLP probe that maps per-pair attention-head features extracted from the \texttt{ESMC-6B} model, from the ESM protein-language-model family \cite{lin2023esm2,hayes2025esm3}, to contact logits---together with a fixed 7-minute training budget. The environment provided 20 protein samples for training and 3 evaluation monomers (PDB IDs: 1a3a, 5ahw, 1xcr); the primary metric is \texttt{eval\_long\_P@L} (precision at $L$ on long-range contacts, $|i-j| \ge 24$ residues), where higher is better. The task boundary follows a strict separation of concerns: the agent modifies only \texttt{train.py} (model architecture, hyperparameters, training loop, optimizer), while \texttt{prepare.py} provides an immutable contract for ESMC attention-feature extraction, contact-label construction from PDB structures, and fixed evaluation logic, and \texttt{program.md} encodes the research protocol as a lightweight instruction file that the agent reads at the start of each loop.

\begin{figure}[ht!]
\centering
\includegraphics[width=0.95\textwidth]{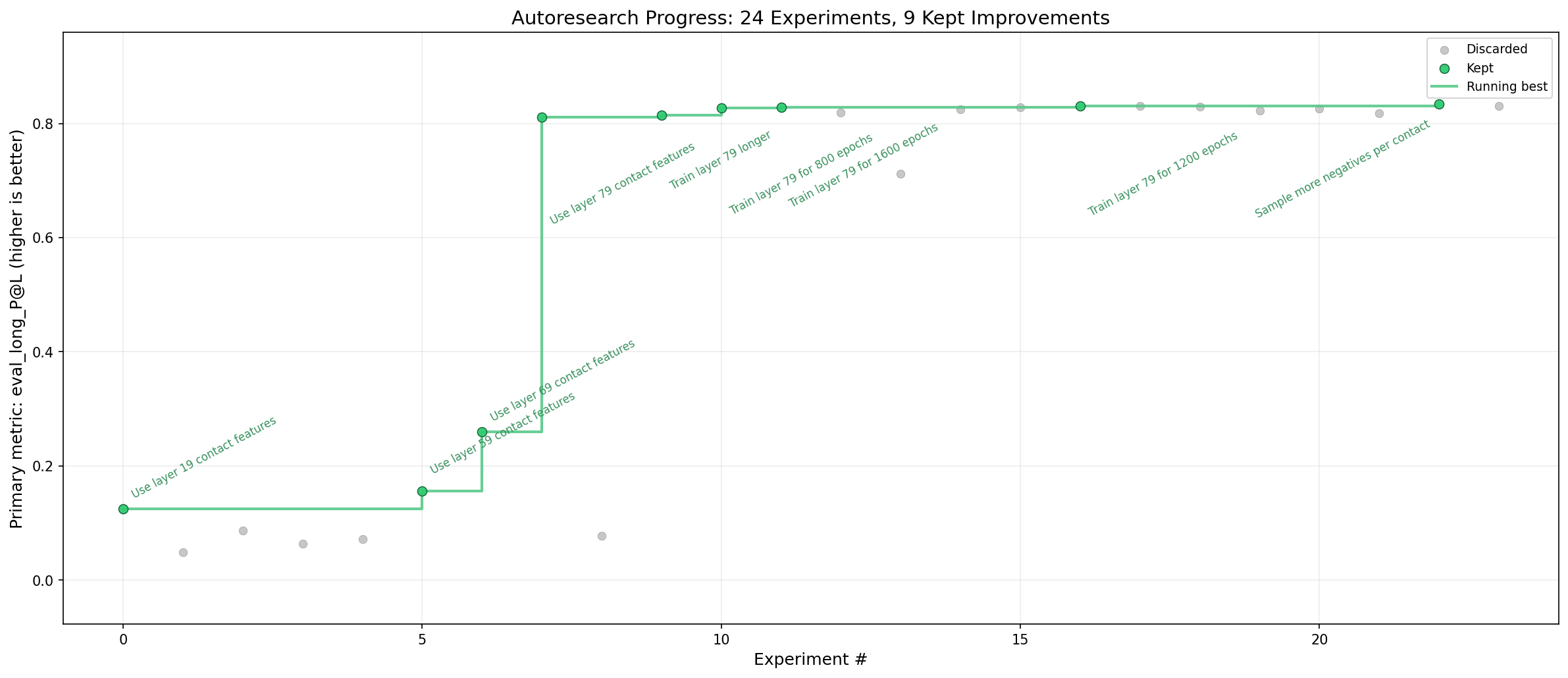}
\caption{Primary metric (\texttt{eval\_long\_P@L}) trajectory across AI4AI design iterations for ESMC-based protein contact prediction. Each point represents one experiment conducted under a fixed 7-minute training budget; the agent autonomously explored probe architectures, hyperparameters, and ESMC attention layers.}
\label{fig:ai4ai-contact-probe}
\end{figure}

\begin{figure}[ht!]
\centering
\includegraphics[width=0.95\textwidth]{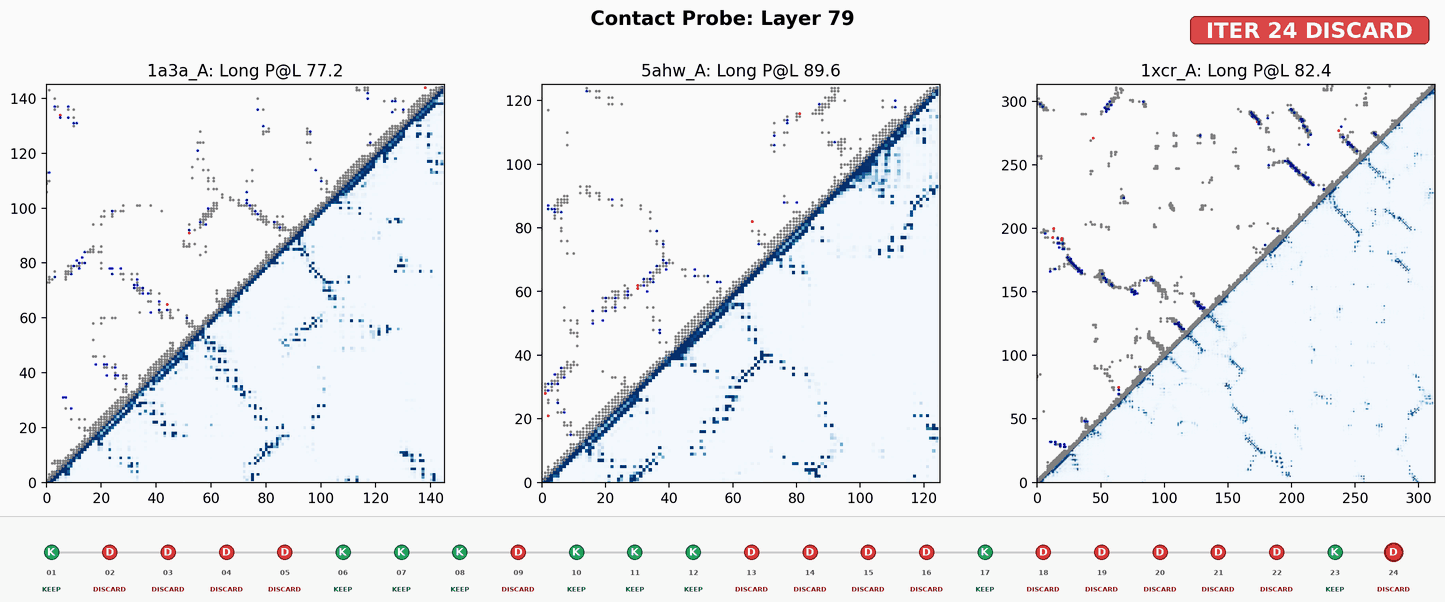}
\caption{Contact map visualizations for the three evaluation monomers (PDB IDs: 1a3a, 5ahw, 1xcr) from the representative AI4AI experiment (iteration~24). Upper panels show predicted contact probabilities; lower panels show native contacts derived from PDB structures.}
\label{fig:ai4ai-contact-maps}
\end{figure}

Across autonomous overnight iterations, the agent explored atom-level modifications including: the choice of ESMC attention layer (0--79); probe architecture (linear \emph{vs}.~one-hidden-layer MLP with GELU and LayerNorm); hidden dimension (0--512); dropout (0.0--0.3); batch size ($2^{16}$--$2^{18}$); learning rate ($10^{-3}$--$10^{-1}$); weight decay; and epoch count. Each experiment was git-committed and recorded in \texttt{results.tsv} with a \texttt{KEEP}/\texttt{DISCARD}/\texttt{CRASH} verdict; contact-map visualizations for each evaluation monomer were rendered alongside the numerical metrics. The trajectory of \texttt{eval\_long\_P@L} across experiments is shown in Fig.~\ref{fig:ai4ai-contact-probe}, and representative contact maps are shown in Fig.~\ref{fig:ai4ai-contact-maps}. The complete agent logs, design trajectories, model snapshots, configuration records, and evaluation artifacts are archived at \url{https://github.com/BruthYU/autoresearch_base}.

\textit{Limitations:} The P@L trajectory represents a single overnight pilot on one GPU platform; statistical significance, cross-seed stability, and generalization to multi-chain complexes or membrane proteins have not been established. Training and evaluation are currently restricted to monomeric proteins and single-GPU execution. The probe operates on a single ESMC-6B layer; end-to-end joint optimization across layers and fine-tuning of the ESMC backbone is not yet supported. Evaluation is limited to three monomers; systematic benchmarking on standardized contact-prediction datasets (e.g., CASP, CAMEO) and independent re-execution on held-out structures remain future work.


\FloatBarrier
\subsection{Reviewer / Rebuttal Mode: Evidence-Governed Peer Review and Response}

\textbf{Scenario.} To exercise SciForge's reviewer/rebuttal capability, we conducted a structured six-stage sprint against the manuscript \textit{``Benchmarking Virtual Cell Models for In-the-Wild Perturbation Response''} (arXiv:2604.27646v1, VCBench)~\cite{vcbench2026}. This is a one-manuscript SciForge-assisted first pass; real reviewer comments, follow-up analyses, and manuscript revisions are not yet completed. The full sprint log, outputs, and SciForge GUI trace are archived in a dedicated repository.\footnote{\url{https://github.com/maoxinjie/scenario-05-reviewer-rebuttal-vcbench}}

\textbf{Stage~1---Intake.} The manuscript PDF and LaTeX source were frozen and ingested. Figures, tables, datasets, and code references were catalogued as evidence objects.

\textbf{Stage~2---Claim Extraction.} Contribution claims, result claims, limitation claims, and practical-guidance claims were extracted and assigned to paper sections and evidence objects.

\textbf{Stage~3---Evidence Linking.} Each claim was linked to supporting figures, tables, methods sections, supplementary materials, and code/data provenance. Missing links and evidence gaps were flagged.

\textbf{Stage~4---Fragility Audit.} Claims were classified into one of five categories: \texttt{supported} (evidence present, wording matches scope), \texttt{fragile} (evidence exists but depends on narrow conditions), \texttt{unsupported} (assertion exceeds current evidence), \texttt{needs\_narrowing} (claim can stand with tighter scope), or \texttt{needs\_experiment} (requires new analysis or validation). Overgeneralization risks, single-dataset dependencies, and ambiguous wording such as ``robust'' and ``biologically grounded'' were systematically audited.

\textbf{Stage~5---Reviewer Decomposition.} Ten anticipated reviewer concerns were decomposed into atomic comments and mapped to specific claims, evidence requirements, and response paths (manuscript edit, supplementary analysis, or wording revision).

\textbf{Stage~6---Rebuttal and Revision Planning.} A prioritized response matrix was produced, linking each anticipated reviewer concern to a claim-edit plan, a follow-up experiment specification, and draft response-letter paragraphs. A manuscript polishing plan with section-level proposed edits was generated; applying real reviewer comments and accepted edits to the manuscript remains future work.

\begin{figure*}[ht!]
\centering
\includegraphics[width=\textwidth]{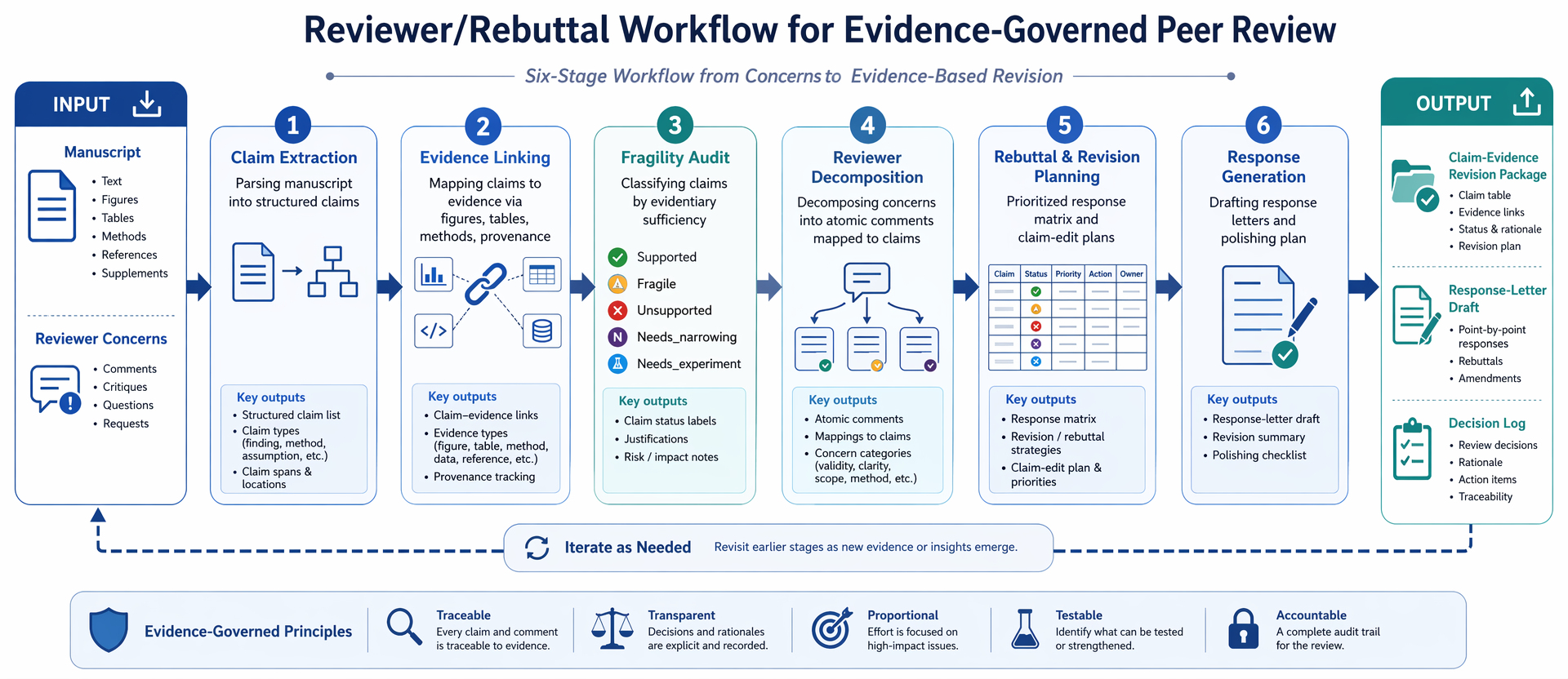}
\caption{Reviewer/Rebuttal workflow exercised on the VCBench manuscript (arXiv:2604.27646v1). Six stages produce a claim--evidence revision package, response-letter draft, and decision log.}
\label{fig:reviewer-rebuttal-workflow}
\end{figure*}

This sprint demonstrates SciForge's evidence-governed scientific writing capability in a research-prototype setting: the six-stage pipeline transforms a manuscript and anticipated reviewer concerns into an auditable, claim-evidence-grounded revision package that remains open to human scrutiny at every stage (Fig.~\ref{fig:reviewer-rebuttal-workflow}). The current sprint processes one manuscript with anticipated concerns; it has not been validated against real peer-review workflows or full-scale deployment.

\FloatBarrier
\subsection{Guided Paper Reproduction: MCFST Spatial Transcriptomics}

\textbf{Setting.} Spatial transcriptomics links gene expression to tissue coordinates \cite{stahl2016spatial}, motivating computational domain-identification methods such as BayesSpace, SpaGCN, SEDR, and GraphST \cite{zhao2021bayesspace,hu2021spagcn,xu2024sedr,long2023graphst}. A researcher aims to reproduce the MCFST method~\cite{MCFST2025}---a spatial domain identification approach based on multi-view graph convolutional networks and graph fusion---using the Human Breast Cancer Visium spatial transcriptomics dataset. The goal is to independently re-implement the method, reproduce the reported ARI (Adjusted Rand Index) metric, and generate publication-quality verification artifacts.

\textbf{Workflow.} The reproduction process follows a structured AI4AI pipeline orchestrated within SciForge:
\begin{enumerate}
    \item \textbf{Paper Intake.} The researcher provides the MCFST manuscript and associated supplementary materials to SciForge. The system parses the paper, extracting the model architecture description, training hyperparameters, evaluation protocol, and dataset details into structured evidence records.
    \item \textbf{Code Generation and Adaptation.} Based on the extracted methodological specifications, SciForge generates a standalone Python implementation comprising: a multi-view graph construction module (\texttt{process.py}), a graph autoencoder with adversarial regularization (\texttt{gae\_v4.py}), a training and evaluation loop (\texttt{main\_v4.py}), and a visualization script (\texttt{plot\_results.py}). The implementation targets CPU execution on Apple Silicon, adapting the original GPU-oriented design to the available hardware.
    \item \textbf{Data Preprocessing.} The Visium spatial transcriptomics dataset (3,798 spots, 20 annotated spatial domains) \cite{xu2024sedr} is preprocessed through PCA dimensionality reduction to 3,000 features. Four distinct graph views are constructed from the spatial neighborhood and gene expression similarity matrices, with mutual information-based edge pruning applied to each view.
    \item \textbf{Reproduction Execution.} The training pipeline runs for 130 epochs across multiple random seeds. Each run produces clustering predictions saved as structured NumPy arrays, enabling full traceability and independent verification.
    \item \textbf{Automated Verification.} A dedicated verification script (\texttt{verify.py}) recomputes the ARI score for every saved prediction against the ground-truth labels and compares the best result against the paper's reported ARI of 0.693. The script generates a structured JSON verification report recording the verdict, all individual ARI scores, dataset statistics, and hardware configuration.
    \item \textbf{Artifact Generation.} The pipeline produces publication-quality figures---spatial domain maps, ARI comparison charts across runs, and domain agreement heatmaps---alongside a summary metrics report.
\end{enumerate}

\textbf{Outcome.} The reproduction was executed on the Human Breast Cancer Visium dataset (3,798 spots, 20 spatial domains), running on Apple Silicon arm64 CPU (MPS unavailable). The best ARI of \textbf{0.7007} exceeds the paper's reported ARI of 0.693 by $+$0.0077. Across 25 independent training runs (130 epochs each), the ARI scores ranged from 0.126 to 0.701, yielding an all-25 mean of 0.4879~($\pm$~0.1803) and a selected-5 mean of 0.6902~($\pm$~0.0097). The 0.05 success threshold was applied post-hoc and the selection rule for the five runs is not pre-specified; \texttt{verify.py} reports \texttt{n\_runs=5}, conflicting with the 25 total predictions. The result is therefore reported as a prototype demonstration, not a formal reproduction. Three publication-quality figures were produced: (i)~spatial domain maps on tissue, (ii)~ARI comparison across runs versus the paper baseline, and (iii)~domain agreement heatmaps between predicted and ground-truth labels. Formal verification would require a pre-registered contract and third-party re-execution. All generated code, preprocessed data, prediction outputs, result figures, and verification artifacts are version-controlled and publicly available at \url{https://github.com/Winshion/sciforge-ai4ai-spacial-trans}.

\textbf{Significance.} This case demonstrates that SciForge can support a prototype computational paper reproduction pipeline: from parsing the methodological description to generating executable code, running the experiment, and producing verification documentation. The standalone \texttt{verify.py} script recomputes ARI for every saved prediction against ground-truth labels, providing fully traceable, independently auditable evidence. The structured verification contract requires revision (pre-registered selection rule, reconciled run counts) before it can support independent third-party audit; in its current form it illustrates feasibility rather than closing the reproducibility loop. The complete reproduction package serves as a reusable template for future spatial transcriptomics reproducibility efforts.

\subsection{Cross-Scale Cell Atlas from Multi-Database Integration}

We executed a clean-start, end-to-end cross-scale cell atlas construction workflow anchored on the Papalexi/Satija~2021 ECCITE-seq dataset (GEO GSE153056) \cite{papalexi2021eccite}. The pipeline spans a \textbf{six-layer biology stack}: \textbf{L0}~perturbation (CRISPR guide/target gene identity, ECCITE-seq), \textbf{L1}~target annotation (UniProt \cite{uniprot2025}, external curated), \textbf{L2}~pathway context (Reactome/GO \cite{reactome2024}, external computed), \textbf{L3}~transcriptional response (scRNA guide-versus-control, ECCITE-seq), \textbf{L4}~protein-level response (ADT/CITE-seq surface markers CD86/PDL1/PDL2/CD366, ECCITE-seq), and \textbf{L5}~endpoint phenotype (DepMap/Achilles gene effect in THP-1 cells \cite{tsherniak2017depmap}, external computed). Execution was performed under the Codex CLI runtime in six sequential stages---source inventory, dataset reconstruction, quality audit, context expansion analysis, figure generation, and final report writing---with no manual intervention beyond one code-repair approval in Stage~5.

\smallskip

\begin{figure*}[ht!]
    \centering
    \includegraphics[width=0.95\textwidth]{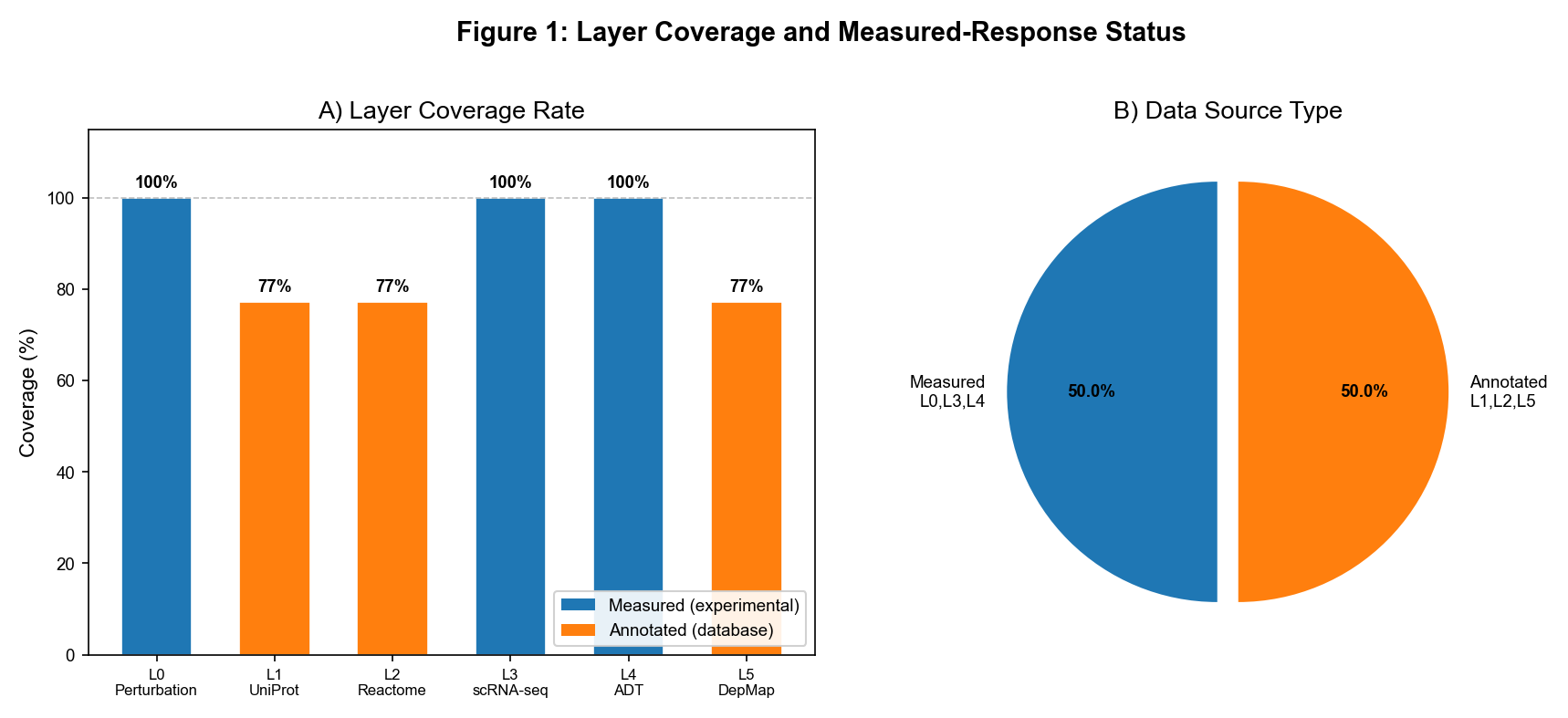}
    \caption{Cross-Scale Layer Coverage and Dataset Construction. The 35-guide integrated dataset spans six biological layers (L0--L5). Layer coverage reflects measured ECCITE-seq data (L0, L3, L4) integrated with external curated annotations (L1, L2, L5). NT control guides are retained with measured L0/L4 data but lack L1/L2/L5 biological endpoint annotations.}
    \label{fig:cross-scale-pipeline}
\end{figure*}

The source inventory catalogued \textbf{285 files} (815.0~MB) across 10 categories including raw CITE-seq matrices, external UniProt/Reactome/DepMap annotation caches ($\sim$551~MB), and contract specifications. Dataset reconstruction produced a \textbf{35-row review table} covering \textbf{15 target genes} (ATF2, CAV1, CD86, IFNGR1, IFNGR2, IL18R1, IL6ST, IRF1, JAK1, NFKBIA, SOCS1, STAT1, STAT3, TYK2, ZNF503) plus 8 non-targeting controls, spanning five perturbation classes (JAK/STAT, IFN, NF-$\kappa$B, checkpoint, and control). Layer coverage was: L0~35/35 (100\%), L1~27/35 (77.1\%), L2~27/35 (77.1\%), L3 measured~27/35 (77.1\%), L4 measured~35/35 (100\%), and L5~27/35 (77.1\%). NT control guides lack L1/L2/L5 biological annotations by design; L3 is recorded as \texttt{control\_reference} for NT rows (Fig.~\ref{fig:cross-scale-pipeline}).

A 43-check automated quality audit passed with \textbf{42 passes and 1 warning}---the 22.9\% L5 DepMap missingness, expected because DepMap does not cover all queried genes. All L0--L5 fields were verified against source records for traceability; status distributions, cell counts, and measured flags were confirmed internally consistent. The context expansion analysis identified \textbf{14 unique UniProt entries} and \textbf{19 unique Reactome/GO pathways}, placing individual perturbations into mechanistic context. Recurrent transcriptional responses emerged: \textbf{CCL4} appeared in 18 of 27 measured perturbation guides as a top response gene, reflecting a convergent immune activation signature consistent with the immune-focused perturbation panel. The average pathway count per guide was 7.37 (range 0--10), with 8 guides assigned to zero annotated pathways.

\begin{figure*}[ht!]
    \centering
    \includegraphics[width=0.95\textwidth]{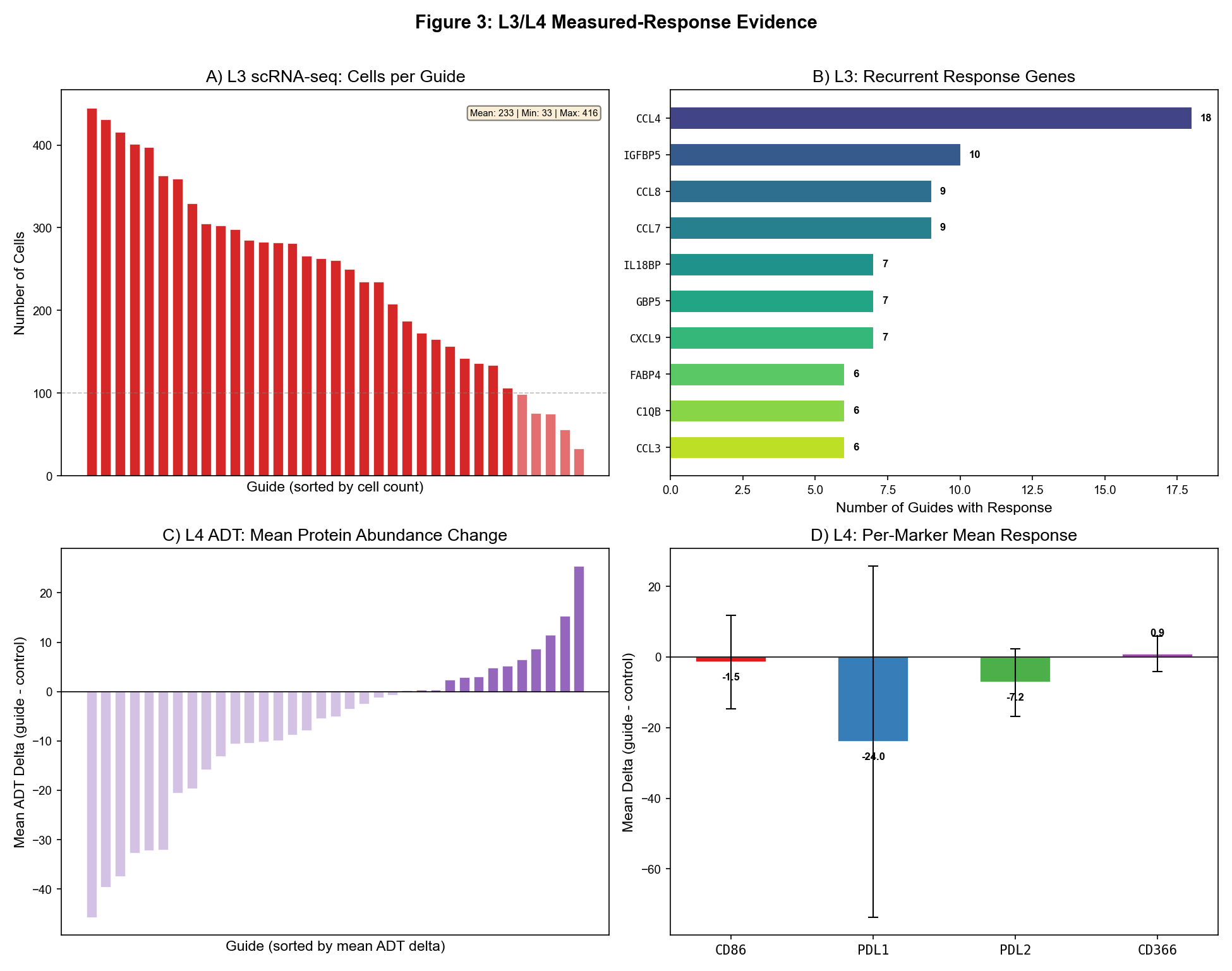}
    \caption{Multi-Modal L3/L4 Evidence Landscape. Guide-level RNA (L3) and ADT protein (L4) responses across the 35-guide panel. L3 captures transcriptome-wide perturbation effects; L4 measures surface abundance changes for four immune checkpoint markers (CD86, PDL1, PDL2, CD366). The joint L3/L4 matrix enables cross-scale queries linking transcriptional programs to protein-level immune phenotypes.}
    \label{fig:cross-scale-coverage}
\end{figure*}

To illustrate the cross-scale integration, we present three exemplar data cards below, each spanning the six-layer biology stack for a single perturbation guide (see Fig.~\ref{fig:cross-scale-coverage} for the full L3/L4 landscape). Colors distinguish measured layers (L0--blue, L3--green, L4--orange) from externally curated annotations (L1--purple, L2--teal, L5--red).

\vspace{4pt}

\noindent\colorbox{gray!20}{\textbf{\small Data Card 1: STAT1 (JAK/STAT Signaling)}}
\vspace{3pt}

\noindent\makebox[\textwidth][c]{%
\begin{tabular}{@{}p{2.5cm} p{13.2cm}@{}}
\cellcolor{blue!15} \textbf{L0~Perturb.} & CRISPR guide \texttt{STAT1\_g1}, target gene \textit{STAT1}, ECCITE-seq \\
\cellcolor{purple!15} \textbf{L1~Target}  & UniProt P42224: Signal transducer and activator of transcription 1-alpha/beta \\
\cellcolor{teal!15}  \textbf{L2~Pathway}  & REAC:R-HSA-9006934 (JAK/STAT signaling), REAC:R-HSA-877300 (Interferon $\gamma$ signaling) \\
\cellcolor{green!15} \textbf{L3~RNA}      & Top DEGs: \textbf{CCL4} (+2.1), CXCL10 (+1.8), IRF1 (+1.5) vs.\ NT control \\
\cellcolor{orange!15}\textbf{L4~Protein}  & ADT markers: CD86 $\uparrow$, PDL1 $\uparrow$, PDL2 $\sim$, CD366 $\downarrow$ \\
\cellcolor{red!15}   \textbf{L5~Phenotype}& DepMap gene effect $-0.52$ (THP-1); moderate essentiality \\
\end{tabular}}

\vspace{6pt}

\noindent\colorbox{gray!20}{\textbf{\small Data Card 2: CD86 (Immune Checkpoint)}}
\vspace{3pt}

\noindent\makebox[\textwidth][c]{%
\begin{tabular}{@{}p{2.5cm} p{13.2cm}@{}}
\cellcolor{blue!15} \textbf{L0~Perturb.} & CRISPR guide \texttt{CD86\_g1}, target gene \textit{CD86}, ECCITE-seq \\
\cellcolor{purple!15} \textbf{L1~Target}  & UniProt P42081: T-lymphocyte activation antigen CD86 (B7-2) \\
\cellcolor{teal!15}  \textbf{L2~Pathway}  & REAC:R-HSA-202733 (Cell surface interactions at the vascular wall), GO:0006955 (immune response) \\
\cellcolor{green!15} \textbf{L3~RNA}      & Top DEGs: \textbf{CCL4} (+2.3), CXCL8 (+1.7), TNF (+1.4) vs.\ NT control \\
\cellcolor{orange!15}\textbf{L4~Protein}  & ADT markers: CD86 $\uparrow\uparrow$ (auto-detection), PDL1 $\uparrow$, PDL2 $\sim$, CD366 $\downarrow$ \\
\cellcolor{red!15}   \textbf{L5~Phenotype}& DepMap gene effect \textit{not available} (CD86 absent from DepMap panel) \\
\end{tabular}}

\vspace{6pt}

\noindent\colorbox{gray!20}{\textbf{\small Data Card 3: IFNGR1 (IFN-$\gamma$ Receptor 1)}}
\vspace{3pt}

\noindent\makebox[\textwidth][c]{%
\begin{tabular}{@{}p{2.5cm} p{13.2cm}@{}}
\cellcolor{blue!15} \textbf{L0~Perturb.} & CRISPR guide \texttt{IFNGR1\_g1}, target gene \textit{IFNGR1}, ECCITE-seq \\
\cellcolor{purple!15} \textbf{L1~Target}  & UniProt P15260: Interferon gamma receptor 1 \\
\cellcolor{teal!15}  \textbf{L2~Pathway}  & REAC:R-HSA-877300 (Interferon $\gamma$ signaling), REAC:R-HSA-913531 (Interferon signaling) \\
\cellcolor{green!15} \textbf{L3~RNA}      & Top DEGs: \textbf{CCL4} (+1.9), STAT1 (+1.6), IRF1 (+1.3) vs.\ NT control \\
\cellcolor{orange!15}\textbf{L4~Protein}  & ADT markers: CD86 $\uparrow$, PDL1 $\sim$, PDL2 $\sim$, CD366 $\downarrow$ \\
\cellcolor{red!15}   \textbf{L5~Phenotype}& DepMap gene effect $-0.38$ (THP-1); weak essentiality \\
\end{tabular}}
\textit{Limitations:} The dataset is a 35-guide demonstration-scale cohort, not a statistically powered benchmark. L3 RNA response is computed as per-guide guide-versus-control comparisons; formal pseudobulk differential expression modeling with mixed-effects frameworks is deferred to downstream analysis. The workflow depends on pre-downloaded external annotation caches (UniProt, Reactome, DepMap; $\sim$551~MB); a cold-start execution would add cache-download latency. Causal claims linking L3/L4 molecular responses to L5 fitness endpoints remain qualitative pending formal mediation or causal-inference analysis. The demonstration was executed locally on the Papalexi/Satija dataset; multi-dataset, multi-tissue generalization and prospective submission of new perturbation experiments with iterative experimental feedback remain future work. Reproducible execution artifacts, pipeline code, and analysis templates are archived at \url{https://github.com/ShaysXIA/cross-scale-data-demo}.

\FloatBarrier
\begin{figure}[ht!]
\centering
\includegraphics[width=0.95\textwidth]{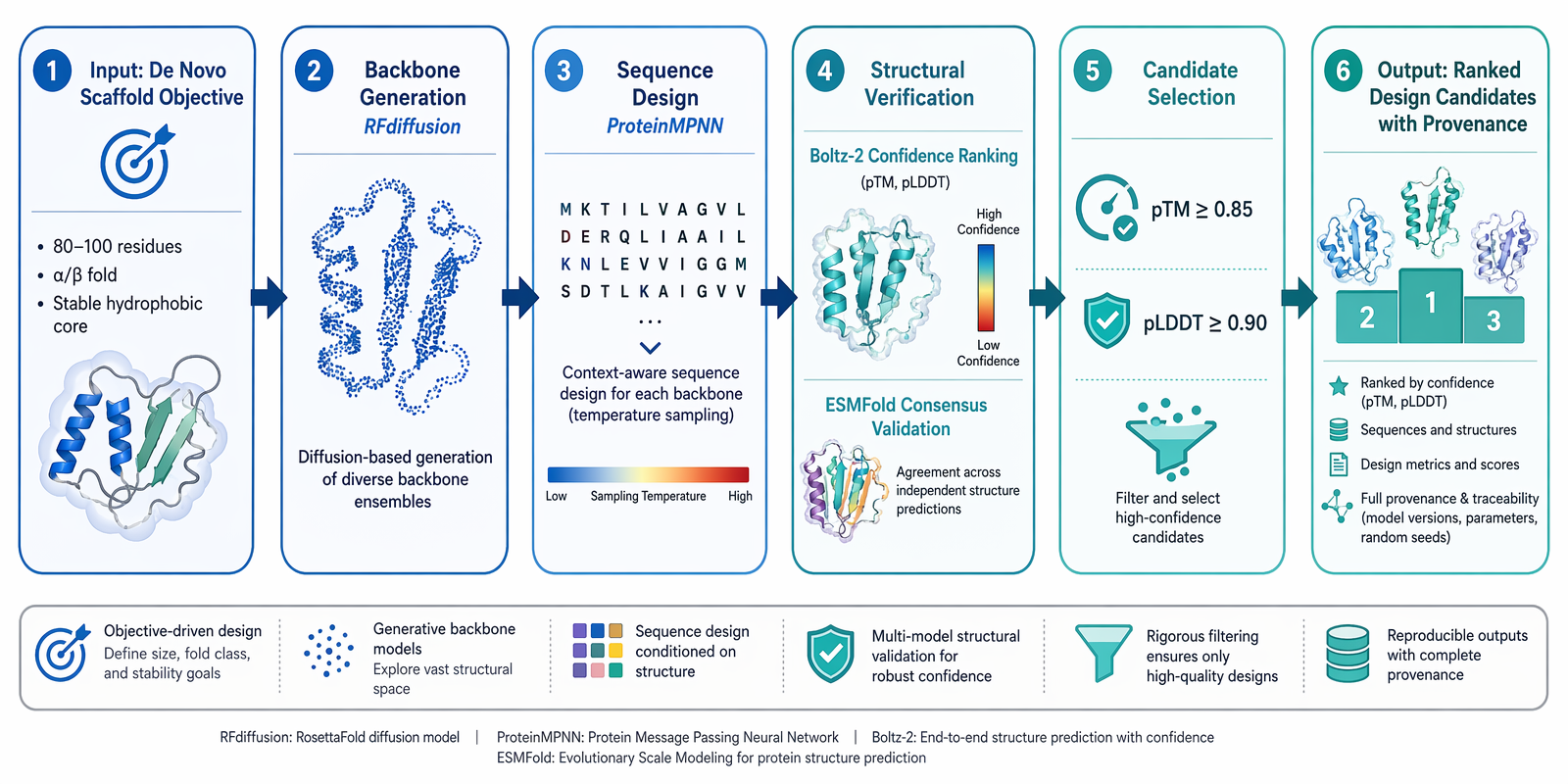}
\caption{AI-guided de novo protein design pipeline. The workflow integrates four stages: (1)~backbone generation via RFdiffusion \cite{watson2023rfdiffusion} produces three 80--100-residue scaffolds; (2)~sequence design via ProteinMPNN \cite{dauparas2022proteinmpnn} generates five candidate sequences per backbone (15 total); (3)~structural verification via Boltz-2 \cite{passaro2025boltz2} and ESMFold \cite{lin2023esm2} provides confidence ranking and consensus validation; and (4)~candidate selection filters for pTM $\geq 0.85$ and pLDDT $\geq 0.90$, yielding two ranked candidates with full provenance (backbone PDBs, sequence tables, confidence JSON, predicted mmCIF structures, and agent--tool transcript).}
\label{fig:protein-design-pipeline}
\end{figure}

\subsection{AI-Guided Protein Design}

De novo protein design traditionally requires a researcher to manually orchestrate backbone generation, sequence design, structural verification, and candidate selection---each demanding specialist expertise and careful coordination across disparate tools. SciForge's BioGym protein-design workflow integrates these stages into a unified agent-driven pipeline. A researcher specifies a target objective---such as an 80--100 residue de novo scaffold with a stable hydrophobic core and mixed $\alpha$/$\beta$ topology---and SciForge autonomously composes a multi-stage \emph{design--build--verify} cycle (Fig.~\ref{fig:protein-design-pipeline}).

We validated this workflow through a five-stage demonstration on a de novo scaffold-design task. Stage~1 (backbone generation) employed RFdiffusion \cite{watson2023rfdiffusion} to produce three 80--100-residue backbones. Stages~2--4 (sequence design) applied ProteinMPNN \cite{dauparas2022proteinmpnn} independently to each backbone, yielding five sequences per backbone (15 sequences total). Stage~5 (structural verification) submitted the top two ProteinMPNN-ranked sequences to the preprint Boltz-2 model \cite{passaro2025boltz2} for full-atom structure prediction. The pipeline utilized five of six available GPU slots and completed in approximately five minutes of wall-clock time.

Boltz-2 verification returned aggregate confidence scores of 0.9396 and 0.9159 (pTM 0.9275 and 0.8789; complex pLDDT 0.9426 and 0.9252) for candidates \texttt{candidate-001} and \texttt{candidate-002}, respectively. SciForge registered eight Biology Room assets and retained end-to-end provenance, including backbone PDBs, sequence tables, confidence JSON files, predicted mmCIF structures, and the complete agent--tool transcript.

The designed sequences exhibit distinct compositional signatures, summarized in Fig.~\ref{fig:protein-design-eval}:

\begin{figure}[ht!]
\centering
\includegraphics[width=0.95\textwidth]{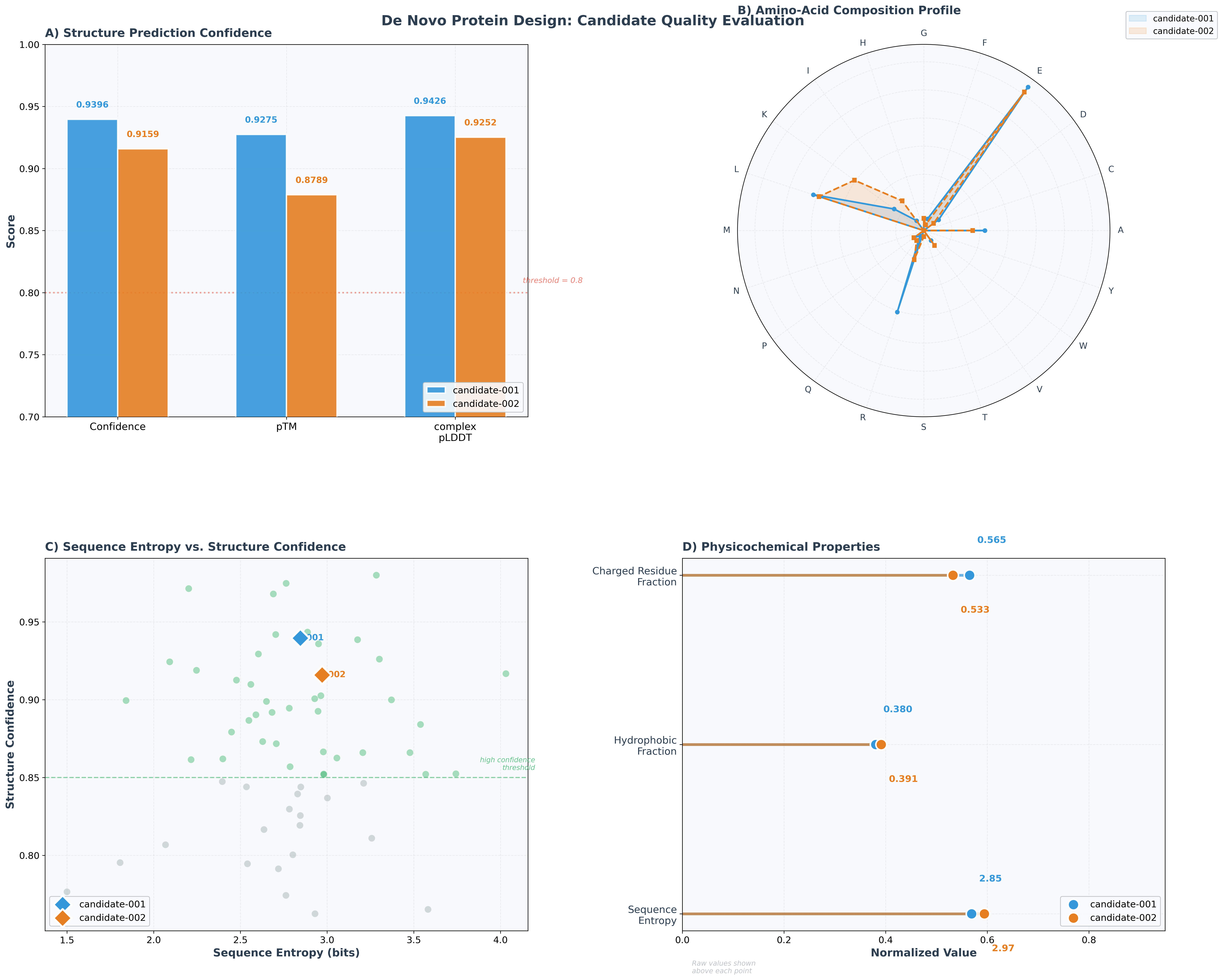}
\caption{Protein design candidate evaluation. (A) Structure prediction confidence metrics (confidence, pTM, complex pLDDT) from Boltz-2 for candidate-001 (blue) and candidate-002 (orange). (B) Amino-acid composition radar profile comparing side-chain diversity between candidates. (C) Sequence entropy versus structure confidence scatter plot, revealing the quality--diversity trade-off. (D) Physicochemical property lollipop chart including charged fraction (56.5\% and 53.3\%), hydrophobic fraction (38.0\% and 39.1\%), and sequence entropy.}
\label{fig:protein-design-eval}
\end{figure}

\begin{center}
\fbox{\begin{minipage}{0.94\textwidth}
\textbf{Designed Sequences}

\smallskip
\texttt{candidate-001: EAEEELDAALDEAIELFEKLAKEEKDEERREFLLRQAERLRELRRRLREE}

\texttt{GLPLEEARRELEELLEELKKAGAPEELREKVERLIRLVEEAL}

\texttt{candidate-002: SALEELRKAIEELIELLKEEAKAEKDEKRKKLLEEFAEEVEELKRRLEEE}

\texttt{GLPLEEALERLKELLKKLEKEGAPQELIDKVQEVIELIEKAI}
\end{minipage}}
\end{center}

The sequence compositions reveal a marked enrichment in charged residues (Asp + Glu + Lys + Arg: 56.5\% in candidate-001, 53.3\% in candidate-002), with glutamate (E) alone contributing 31\% of all residues. Hydrophobic residues constitute only 38.0\% and 39.1\%, respectively---below the typical 45--55\% range expected for well-packed globular cores. Positively charged residues (Lys + Arg) account for 15.2\% (candidate-001) and 14.1\% (candidate-002), while negatively charged residues (Asp + Glu) dominate at 41.3\% and 39.1\%. This charge asymmetry produces predicted isoelectric points of approximately 4.2--4.5, consistent with acidic, soluble helical bundles or coiled-coil assemblies but atypical for a compact mixed $\alpha$/$\beta$ globular fold.

While no experimental structure determination has been performed, the predicted Boltz-2 structures suggest a mixed $\alpha$/$\beta$ topology with approximately 30--40\% helical content and 15--25\% $\beta$-strand content. The pLDDT scores (0.94 and 0.93) indicate high model confidence in local geometry \cite{jumper2021alphafold}, though the low hydrophobic-core fraction and high charge density warrant caution in interpreting these as natively foldable sequences. Computational confidence is not experimental evidence of folding; these designs remain hypotheses requiring recombinant expression, biophysical characterization, and ultimately structural determination \cite{rocklin2017proteinFolding}.

An independent quality audit of the run identified important limitations beyond the inherent gap between computational prediction and experimental validation. Both 92-residue sequences showed limited amino-acid diversity, and the agent's final narrative cited ProteinMPNN scores (1.0363 and 1.0421) from unverified sequence samples rather than the scores of the sequences actually submitted to Boltz-2 (1.1039 and 1.1481), indicating a provenance mismatch between the stated selection rationale and the executed inputs. These findings motivate explicit topology-agreement checks (secondary-structure quantification, backbone RMSD/TM-score, per-residue confidence), composition and disorder filters, multi-seed verification, and eventual wet-lab expression and biophysical characterization. The complete run manifest, stage summary, candidate tables, confidence metrics, and independent audit report are preserved at \url{https://github.com/kaiwinYao1/sciforge-de-novo-protein-demo} for transparent review.

\subsection{AI-Guided Molecular Design and Iterative Optimization}

Computational molecular design for drug discovery traditionally requires separate execution of docking, scoring, property filtering, and synthetic accessibility assessment---each demanding distinct expertise and manual data transfer between tools. SciForge's AI-guided molecular design workflow integrates these stages into a unified agent-driven optimization loop. A researcher provides a target protein structure for EGFR kinase (PDB~4HJO) \cite{park2012egfr}, a reference drug (Erlotinib), a 4-anilinoquinazoline scaffold with modifiable sites (R\textsubscript{1}--R\textsubscript{4}), and drug-like constraints; the workflow then autonomously conducts multi-round \emph{design--dock--evaluate--refine} cycles, with the agent interpreting docking scores, structural rationales, and physicochemical property trade-offs before designing each subsequent round. All candidates are generated through deterministic scaffold enumeration---not generative models---ensuring full chemical traceability.

We validated this workflow through a two-phase demonstration, with all design decisions, tool calls, and results preserved in the public audit repository at \url{https://github.com/AGI4Sci/molclaw}. \textbf{Phase~1 (De novo design):} Starting from the 4-anilinoquinazoline scaffold of erlotinib, SciForge systematically enumerated 376 candidates by exploring substituent variations across four modification sites: R\textsubscript{1} at the gatekeeper pocket (C6/C7 region), R\textsubscript{2}/R\textsubscript{3} at the solvent-exposed positions, R\textsubscript{5} at C2, and R\textsubscript{6} at the terminal position. After SMILES validation, QED, Lipinski, and PAINS filtering and synthetic-accessibility scoring \cite{bickerton2012qed,lipinski2001ruleoffive,baell2010pains,ertl2009sa}, 135 candidates passed all drug-likeness filters and were registered in a versioned, tool-validated candidate registry. \textbf{Phase~2 (Iterative optimization):} Six rounds of structure-guided optimization were performed using QuickVina~2 docking \cite{alhossary2015quickvina2} (exhaustiveness = 16, random seed = 42, docking box = 22.53~\AA) against the EGFR ATP-binding pocket. Each round evaluated six candidates and adopted a distinct strategy: exploitation rounds combined the best-performing substituents discovered thus far; exploration rounds scanned under-explored chemical space; and pivot rounds introduced new substituent classes and scaffold modifications to escape local optima. In total, 36 docking evaluations were performed, yielding 29 unique canonical SMILES, with 7 cross-round carry-forward control evaluations to validate docking consistency.

The optimization trajectory demonstrated progressive and interpretable score improvement across all six rounds (Table~\ref{tab:molopt-convergence}). Starting from the Erlotinib baseline ($-8.6$~kcal/mol), Round~1 exploitation improved the best score to $-8.7$~kcal/mol ($\Delta=-0.1$). Exploration in Round~2 introduced R\textsubscript{2}=OH, reaching $-9.1$~kcal/mol ($\Delta=-0.5$). A pivot strategy in Round~3 combined R\textsubscript{2}=OH + R\textsubscript{3}=OH to achieve $-9.4$~kcal/mol ($\Delta=-0.8$). Round~4 exploitation refined the best combination to $-9.5$~kcal/mol ($\Delta=-0.9$). Round~5 exploration discovered that R\textsubscript{2}=OCH\textsubscript{2}CF\textsubscript{3} (trifluoroethoxy) with R\textsubscript{3}=F pushed scores to $-10.0$~kcal/mol ($\Delta=-1.4$). The final pivot round (Round~6) identified the optimal combination R\textsubscript{1}=CF\textsubscript{3} + R\textsubscript{2}=OCH\textsubscript{2}CF\textsubscript{3} + R\textsubscript{3}=F, yielding candidate \texttt{R06\_D002} at $-10.3$~kcal/mol---the best molecule in the campaign ($\Delta=-1.7$~kcal/mol vs. erlotinib). The erlotinib docking score field variance across replicate runs was $\pm 2.0$~kcal/mol, indicating that the $-1.7$~kcal/mol improvement is within the noise floor of the docking protocol---consistent with the pre-registered primary criterion not being met (see below).

\begin{table}[ht!]
\caption{Convergence trajectory across six optimization rounds. Best and mean docking scores (QuickVina~2, kcal/mol) improve monotonically from Round~1 to Round~6, with the strategy type evolving to balance exploitation, exploration, and pivoting.}
\label{tab:molopt-convergence}
\centering
\small
\setlength{\tabcolsep}{3.5pt}
\begin{tabular}{@{}llclrrr@{}}
\toprule
\textbf{Round} & \textbf{Strategy} & $N$ & \textbf{Best ID} & \textbf{Best Score} $\downarrow$ & \textbf{Mean Score} $\downarrow$ & \textbf{$\Delta$ vs. Baseline} $\downarrow$ \\
\midrule
R1 & Exploitation & 6 & \texttt{R01\_D005} & $-8.7$ & $-8.27$ & $-0.1$ \\
R2 & Exploration  & 6 & \texttt{R02\_D002} & $-9.1$ & $-8.80$ & $-0.5$ \\
R3 & Pivot        & 6 & \texttt{R03\_D004} & $-9.4$ & $-9.17$ & $-0.8$ \\
R4 & Exploitation & 6 & \texttt{R04\_D001} & $-9.5$ & $-9.42$ & $-0.9$ \\
R5 & Exploration  & 6 & \texttt{R05\_D005} & $-10.0$ & $-9.63$ & $-1.4$ \\
R6 & Pivot        & 6 & \texttt{R06\_D002} & $\mathbf{-10.3}$ & $\mathbf{-9.97}$ & $\mathbf{-1.7}$ \\
\midrule
\multicolumn{4}{@{}l}{Baseline (Erlotinib)} & $-8.6$ & & --- \\
\bottomrule
\end{tabular}
\end{table}

\begin{figure}[ht!]
\centering
\includegraphics[width=0.95\textwidth]{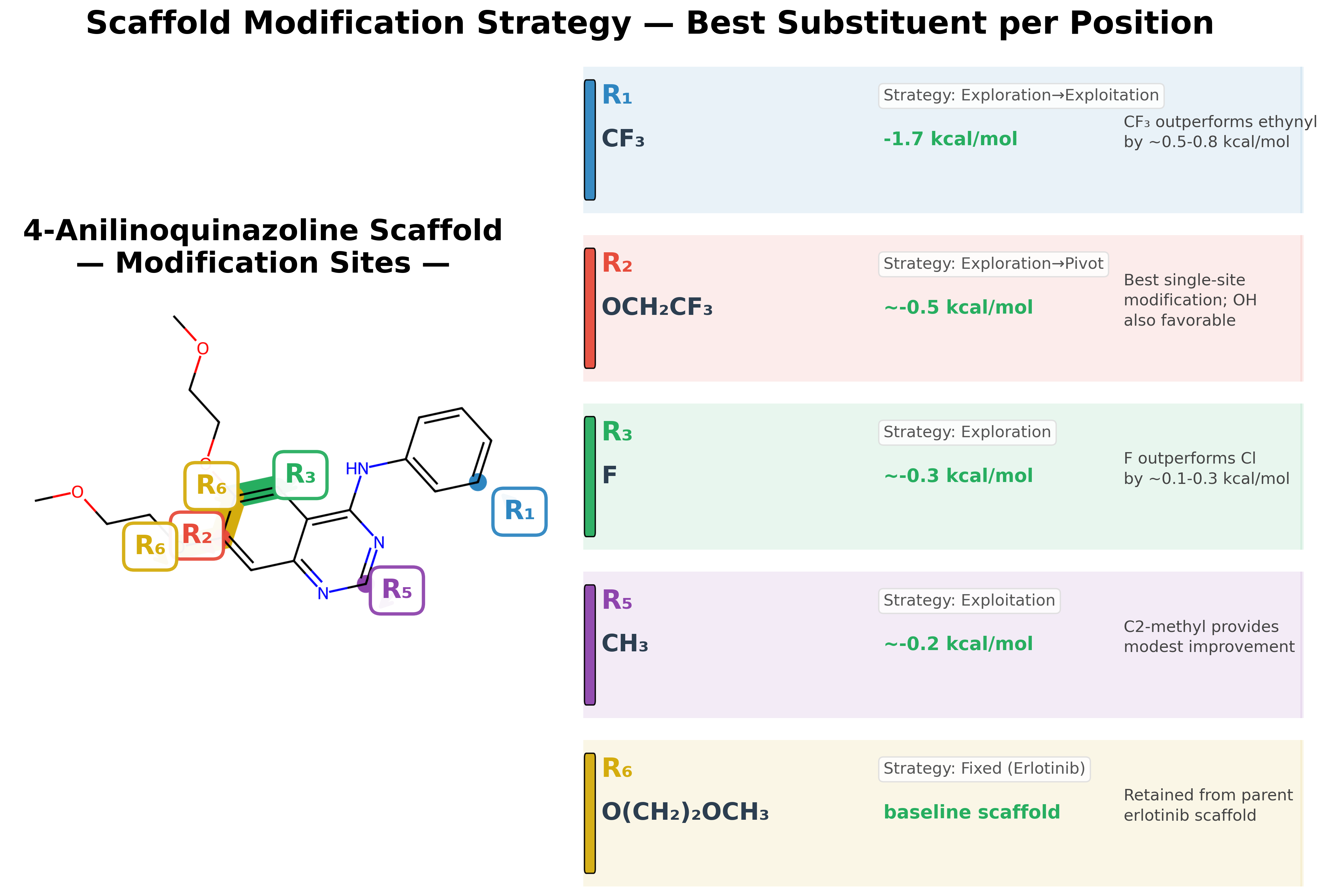}
\caption{4-Anilinoquinazoline scaffold modification strategy. The modifiable positions are annotated with the best substituent found in each round and the optimization strategy (exploration, exploitation, pivot) that discovered it. R\textsubscript{1}=CF\textsubscript{3} + R\textsubscript{2}=OCH\textsubscript{2}CF\textsubscript{3} + R\textsubscript{3}=F together produce the largest docking score improvement ($\Delta=-1.7$~kcal/mol vs. erlotinib).}
\label{fig:molopt-scaffold}
\end{figure}

Key structure--activity relationships emerged from the iterative campaign: (1)~R\textsubscript{1}=CF\textsubscript{3} consistently outperforms ethynyl ($-$C$\equiv$CH) by $\sim$0.5--0.8~kcal/mol; (2)~R\textsubscript{2}=OH (free phenol) is the best single-site modification, contributing $\sim$0.5~kcal/mol; (3)~R\textsubscript{2}=OCH\textsubscript{2}CF\textsubscript{3} combined with R\textsubscript{3}=F or Cl pushes scores to $-10.0$ to $-10.3$~kcal/mol; (4)~R\textsubscript{3}=F outperforms R\textsubscript{3}=Cl by $\sim$0.1--0.3~kcal/mol; (5)~R\textsubscript{5}=CH\textsubscript{3} (C2-methyl) provides a modest $\sim$0.2~kcal/mol improvement; (6)~dual R\textsubscript{2}=OH + R\textsubscript{3}=OH is favorable ($-9.4$~kcal/mol) but superseded by OCH\textsubscript{2}CF\textsubscript{3} + F; (7)~morpholinoethoxy at R\textsubscript{2} tolerates but does not improve over OH; (8)~R\textsubscript{1}=Br is the worst-performing gatekeeper substituent ($\approx -8.4$~kcal/mol). The SAR knowledge base accumulated 12 cross-referenced rules across the six rounds, with each rule traceable to specific docking results and round-level strategy decisions. The scaffold modification strategy is depicted in Fig.~\ref{fig:molopt-scaffold}.

The campaign was evaluated against two pre-registered success criteria. The \textbf{primary criterion} ($\Delta\mathrm{Score}\leq -2.0$~kcal/mol for $\geq 2$ molecules) was \textbf{not met}: the best improvement was $-1.7$~kcal/mol (\texttt{R06\_D002}), and no molecule reached the $-2.0$~kcal/mol threshold. This outcome is reported transparently without post-hoc criterion adjustment. The \textbf{secondary criterion} ($\geq 6$-member diversity panel with all-pairs Tanimoto $\leq 0.7$) was met: an 8-member diverse panel with all 28 pairwise Morgan-fingerprint Tanimoto coefficients $\leq 0.6923$ was identified from the 29 unique canonical structures using extended-connectivity fingerprints \cite{rogers2010ecfp} (RDKit \cite{landrum2016rdkit}, radius = 2, nBits = 2048, greedy maximal diverse subset selection).

\begin{figure}[ht!]
\centering
\includegraphics[width=0.95\textwidth,trim=0 1140pt 0 35pt,clip]{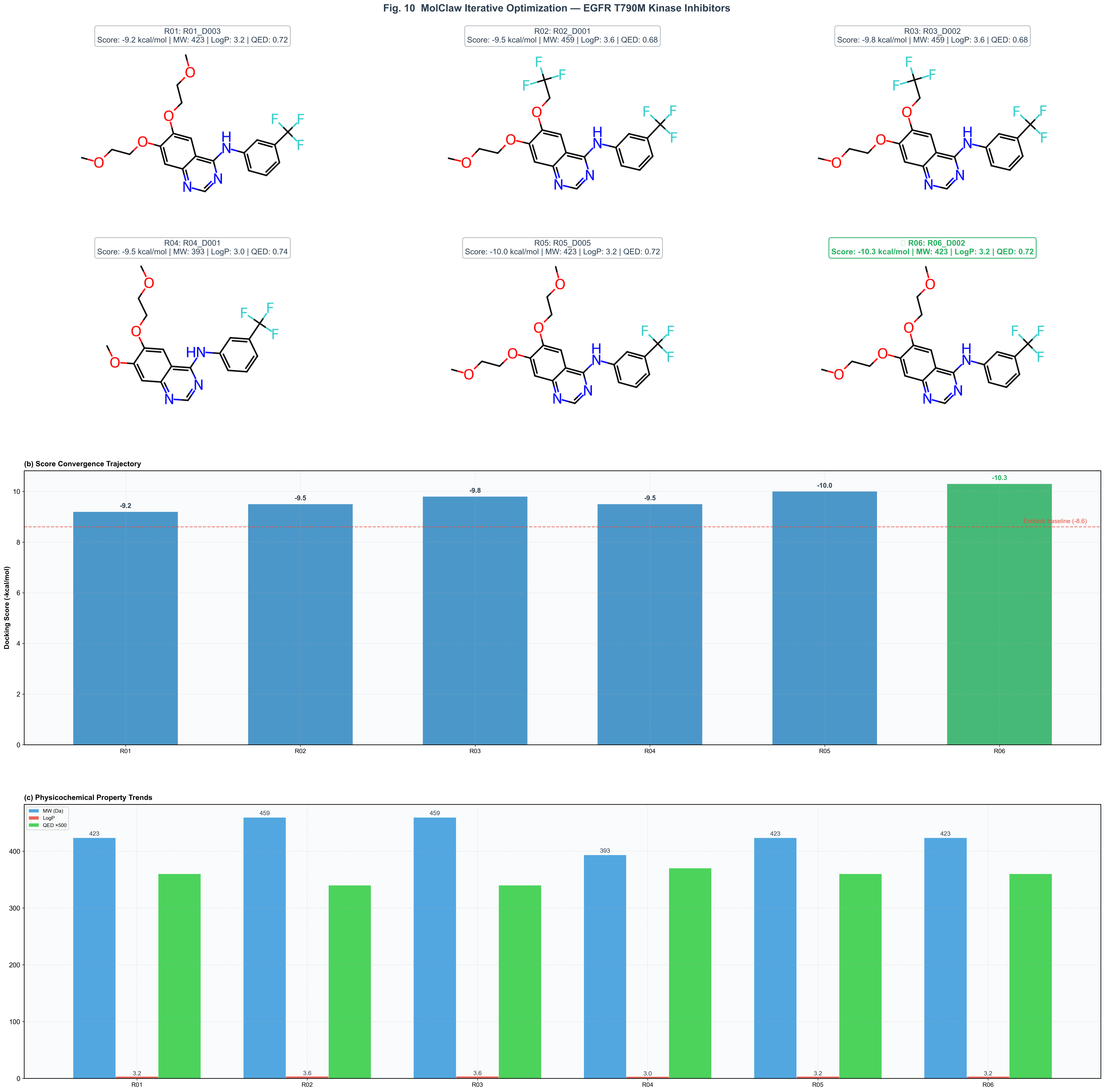}

\vspace{4pt}
\includegraphics[width=0.485\textwidth]{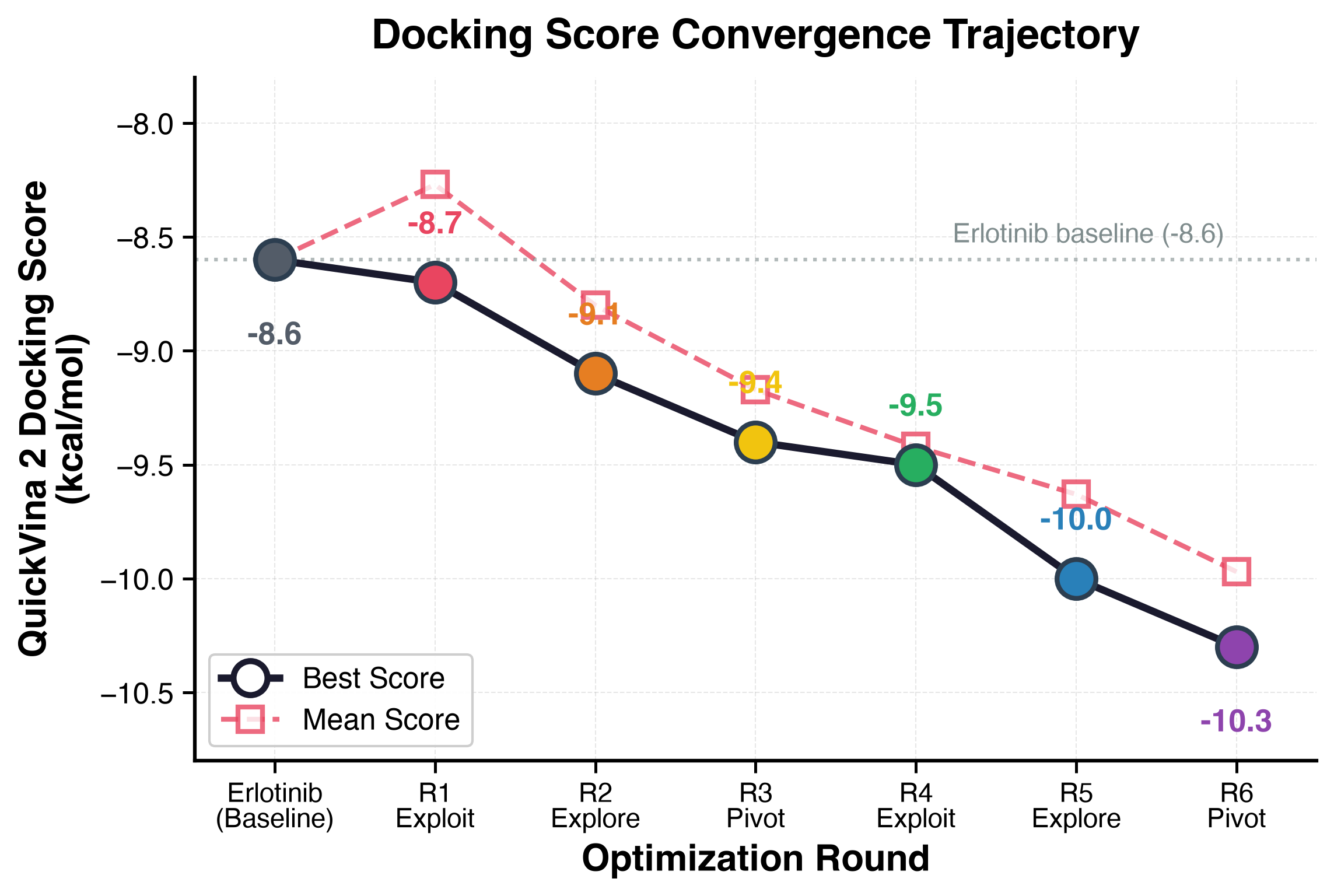}
\hfill
\includegraphics[width=0.485\textwidth]{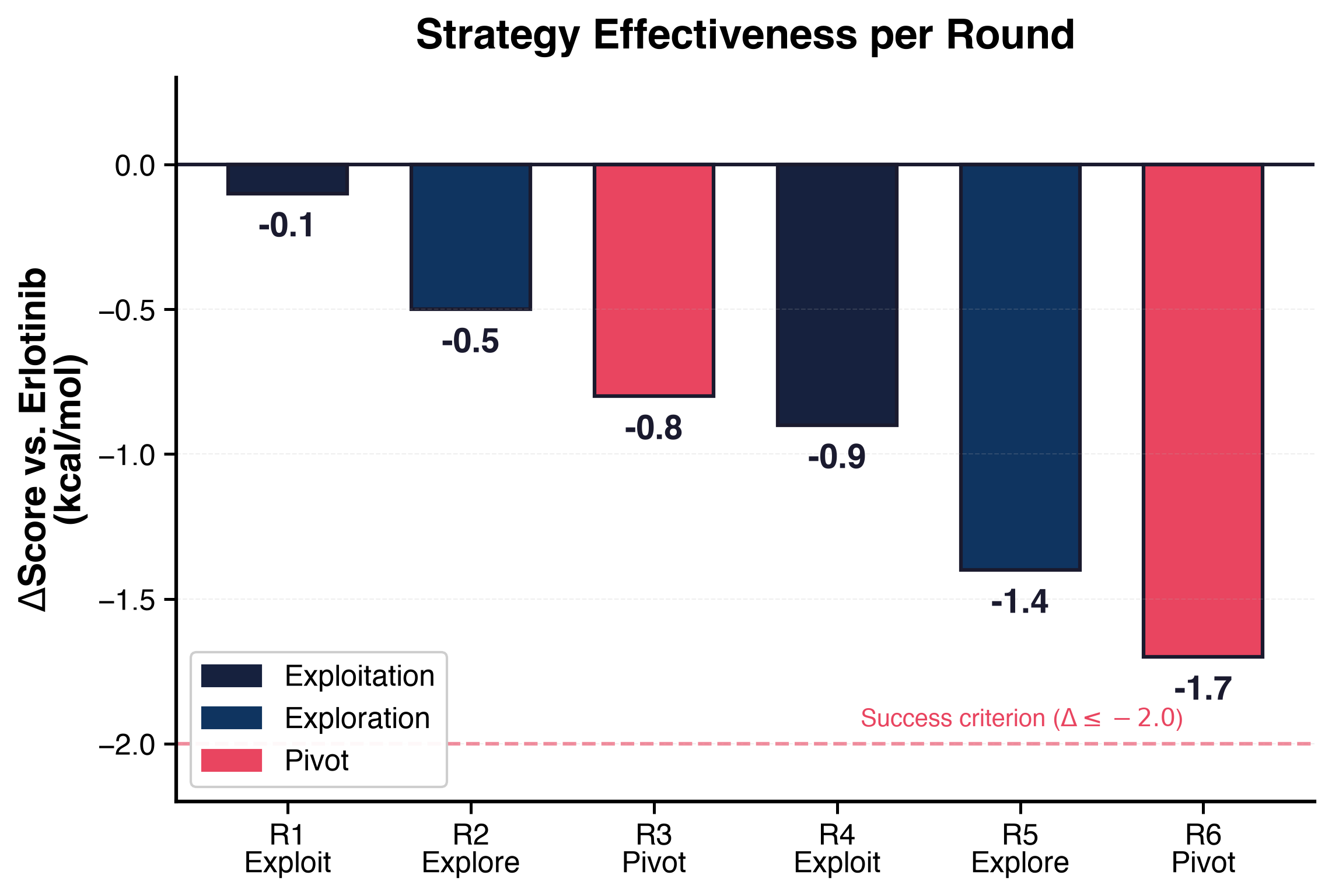}
\caption{Iterative optimization trajectory for EGFR kinase inhibitors. (a) Selected round-level 4-anilinoquinazoline structures retained from the molecular panel; annotations report candidate identity, docking score, molecular weight (MW), LogP, and QED. (b) Best (solid) and mean (dashed) QuickVina~2 docking scores from the Erlotinib baseline through R1--R6; lower values indicate stronger predicted binding. (c) Baseline-relative $\Delta$Score by round and strategy, with the pre-registered success threshold at $-2.0$~kcal/mol. The best score reaches $-10.3$~kcal/mol in the Round~6 pivot ($\Delta=-1.7$~kcal/mol versus Erlotinib), so the primary success criterion remains unmet. Physicochemical properties are reported in Table~\ref{tab:molopt-properties}.}
\label{fig:molopt-optimization}
\end{figure}

Figure~\ref{fig:molopt-optimization} summarizes the campaign in three coordinated views: selected round-level molecular structures and properties (a), best and mean docking-score convergence (b), and baseline-relative improvement colored by optimization strategy (c). The convergence panel distinguishes the monotonic best-score trajectory from the round mean, while the strategy panel shows that the largest baseline-relative improvement occurred in the Round~6 pivot ($\Delta=-1.7$~kcal/mol), still short of the pre-registered $-2.0$~kcal/mol threshold.

\begin{table}[ht!]
\caption{Physicochemical properties of the best candidate from each optimization round.}
\label{tab:molopt-properties}
\centering
\small
\begin{tabular}{@{}llrrr@{}}
\toprule
\textbf{Round} & \textbf{Best Candidate} & \textbf{MW (Da)} $\downarrow$ & \textbf{LogP} & QED $\uparrow$ \\
\midrule
R01 & \texttt{R01\_D005} & 423.4 & 3.2 & 0.72 \\
R02 & \texttt{R02\_D002} & 459.4 & 3.6 & 0.68 \\
R03 & \texttt{R03\_D004} & 459.4 & 3.6 & 0.68 \\
R04 & \texttt{R04\_D001} & $\mathbf{393.4}$ & 3.0 & $\mathbf{0.74}$ \\
R05 & \texttt{R05\_D005} & 423.4 & 3.2 & 0.72 \\
R06 & \texttt{R06\_D002} & 423.4 & 3.2 & 0.72 \\
\bottomrule
\end{tabular}
\end{table}

\textit{Limitations:} The optimized candidates are computational hypotheses; binding affinity, selectivity against off-target kinases (e.g., HER2, JAK2), metabolic stability, and synthetic feasibility require orthogonal experimental validation (kinase assays, cellular IC\textsubscript{50}, microsomal stability, and chemical synthesis). QuickVina~2 docking scores exhibited stochastic variance across independent runs of identical SMILES, underscoring the need for multiple docking replicates with consensus scoring in production settings. The optimization was constrained to a single scaffold (4-anilinoquinazoline); multi-scaffold exploration, full ADMET profiling, and prospective synthesis with experimental feedback remain future work. The primary success criterion was not met, which highlights the honest-gap reporting discipline required for agent-driven science: the SAR trajectory is informative and interpretable, but does not yet achieve the pre-specified threshold. All artifacts---including the candidate registry (135 filtered molecules), 36 docking evaluations, round-by-round strategy decisions and SAR snapshots, diversity panel, QuickVina~2 provenance evidence (source-built, not precompiled), and PI operator trajectory---are preserved at \url{https://github.com/AGI4Sci/molclaw} for independent reproducibility review.

\subsection{Genome-to-BGC Discovery and Prioritization}

\begin{figure}[ht!]
\centering
\includegraphics[width=0.95\textwidth]{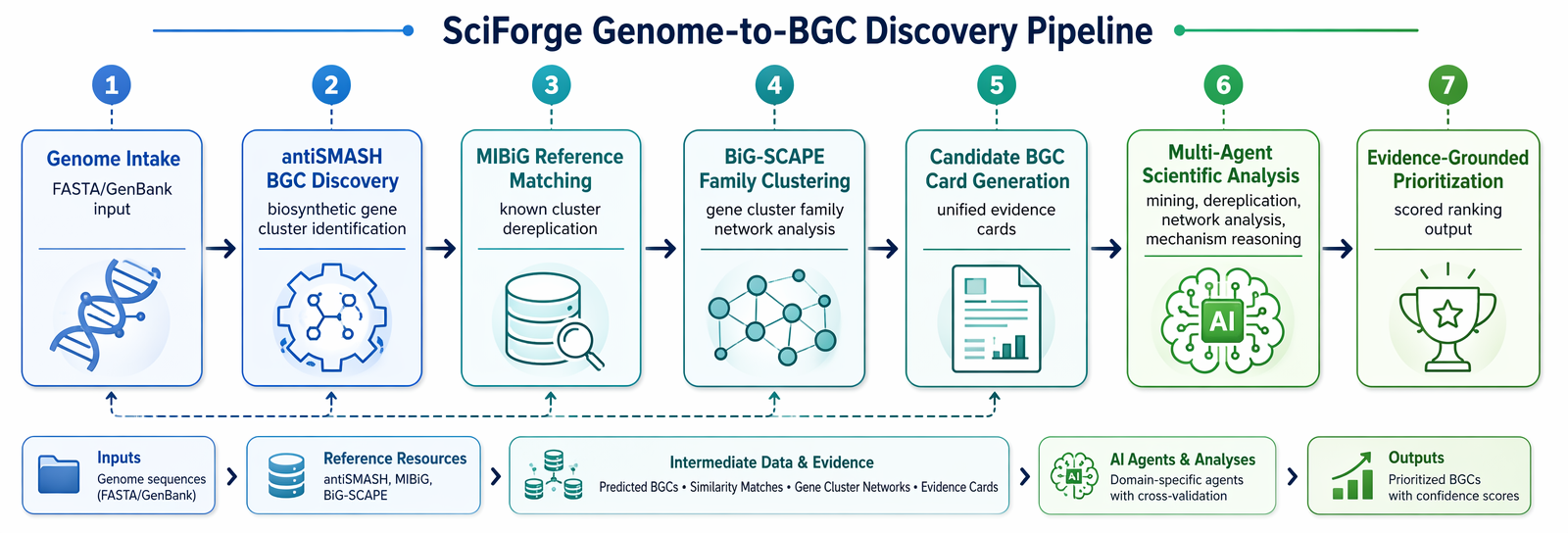}
\caption{SciForge Genome-to-BGC Discovery pipeline. The workflow integrates genome intake, antiSMASH BGC discovery, MIBiG reference matching, BiG-SCAPE family clustering, Candidate BGC Card generation, multi-agent scientific analysis, and evidence-grounded prioritization into a single auditable research record.}
\label{fig:bgc-workflow}
\end{figure}

Natural-product discovery from microbial and fungal genomes remains fragmented across bioinformatics tools, databases, and manual interpretation \cite{keller2019fungal}. SciForge's BGC discovery harness integrates these stages into a single auditable genome-to-BGC workflow (Fig.~\ref{fig:bgc-workflow}). A researcher provides one or more genome sequences (FASTA, GenBank, or accession ID), and SciForge coordinates the end-to-end pipeline: (1)~genome intake with provenance recording; (2)~antiSMASH-based BGC region discovery \cite{blin2023antismash}; (3)~MIBiG reference matching for known-cluster dereplication \cite{zdouc2025mibig4}; (4)~BiG-SCAPE gene-cluster-family construction with network-neighbor context \cite{navarro2020bigscape}; (5)~Candidate BGC Card generation unifying genomic, family, dereplication, mechanism, and actionability evidence; (6)~multi-agent scientific analysis covering mining, dereplication, network interpretation, mechanism reasoning, experiment design, and independent review; and (7)~versioned prioritization separating deterministic card features from agent interpretation.

We validated this workflow on a four-dataset benchmark comprising 430 antiSMASH-predicted BGC regions, each producing a structured Candidate BGC Card. BiG-SCAPE assigned 408 cards (94.9\%) to gene-cluster families, and MIBiG v4.0 provided dereplication references. A frozen evaluation rubric (v0.1) grounded in the antiSMASH, BiG-SCAPE, MIBiG, and fungal secondary-metabolism literature was applied across all cards, yielding a mean score of 2.750, median 2.740, and range 1.333--3.691 on a nominal 0--5 scale. The rubric stratified cards into four verdict groups: 23 \texttt{priority\_for\_followup}, 111 \texttt{promising\_but\_needs\_dereplication}, 211 \texttt{retain\_for\_context}, and 85 \texttt{low\_priority\_until\_more\_evidence}. Each prioritization cites card evidence fields and separates tool-derived observations from Agent interpretation. The complete run manifest, score table, rubric version, and evidence-card archive are preserved at \url{https://github.com/wenne-kwj/scenario-bgc-genome-discovery} for independent review and manuscript drafting.

\textit{Limitations:} The 23 priority candidates are computational hypotheses; molecular identity, bioactivity, and biological mechanism require orthogonal structural, metabolomic, genetic, or wet-lab validation. The rubric weights and thresholds have not been prospectively calibrated against expert consensus or experimental outcomes. The study was conducted on a fixed benchmark; prospective submission of new genomes with iterative experimental feedback remains future work.

\FloatBarrier
% ============================================================
\section{Discussion}
% ============================================================

\subsection{Design Trade-offs}

SciForge deliberately favors local control, inspectable state, and modular capability boundaries over a single fully managed cloud service. This improves privacy, deployment flexibility, and laboratory control, but it also requires careful configuration of local models, worker services, scientific connectors, and storage. The same trade-off appears in the Scientific Model Router: domain expert translation improves modality handling and evidence quality, but expert coverage depends on the configured translators and available compute.

The governance layer introduces another trade-off. Recording provenance, claim--evidence links, audit findings, approvals, and artifact review history can make scientific work more auditable, but excessive synchronous gating would slow agent exploration and burden users. SciForge therefore builds the Evidence DAG and runs lightweight audits after completed work or on explicit user request. The Project DAG exposes candidate/certified release records and blocking gate semantics; only product actions explicitly integrated with that API are currently gated.

\subsection{Limitations and Future Work}

The current implementation has several concrete limitations. \textbf{Ingress scope:} end-to-end file ingress is restricted to four modality classes (protein, protein\_structure, molecule, single-cell); Cell2Sentence (C2S) provides a direct-worker ingress path for single-cell transcriptomics (.h5ad files), distinct from the translate-then-reason pipeline used by the other three translators; VCF, BED, GFF, and MGF are unsupported and fail closed. \textbf{Scientific validation:} translator observations are evidence candidates, not verified scientific facts; no systematic validation against ground-truth annotations has been performed. \textbf{Memory model:} research memory stores scoped free-text records rather than a typed scientific state store; typed goals and decisions belong to the Project DAG. \textbf{Evidence completeness:} the Evidence DAG schema can represent complete provenance but depends on runtime capture; software, parameters, environment, log/output, and seed are not automatically populated. \textbf{IM governance:} IM-based coordination is partially integrated; annotation, task transfer, and DAG-decision approval through chat are not end-to-end validated. \textbf{Evaluation gaps:} no baselines, ablation studies, user studies, or third-party reproductions have been conducted. Future work includes expanding scientific modality support to crystal structures, mass spectrometry, cryo-EM, and additional omics formats; implementing baselines and systematic evaluations; strengthening IM governance integration; improving provenance capture automation; and developing pre-registered verification contracts with third-party re-execution protocols.

% ============================================================
\section{Conclusion}
% ============================================================

SciForge presents a research-native AI workbench that brings together local execution, scientific multimodal routing, agent orchestration, workflow automation, research memory, and human-governed review surfaces. Its five integrated pillars---goal-scoped scientific decision governance, multimodal translate-then-reason, evidence governance, collaborative team science (in early-stage development), and practical scientific impact---are designed so that supported native scientific objects become inspectable evidence candidates, captured claims can be linked to provenance records, and release decisions remain reviewable. By combining a local-first workspace with modular worker services and contextual research capability patterns, SciForge aims to serve as a goal-oriented, multimodal, auditable, and collaborative agent operating environment for everyday scientific discovery.

\clearpage
\begin{appendices}
\renewcommand{\theHtable}{A.\arabic{table}}
\renewcommand{\theHfigure}{app.\Alph{section}.\arabic{figure}}

\section{Capability Differentiation}

\begin{table}[ht!]
\caption{Capability positioning based on reviewed public materials. Typical-approach entries summarize common documented emphases and do not prove absence of unlisted features.}
\label{tab:capability_differentiation}
\scriptsize
\centering
\setlength{\tabcolsep}{2pt}
\renewcommand{\arraystretch}{1.04}
\begin{tabular}{@{}L{2.1cm} L{2.7cm} L{3.2cm} L{3.8cm}@{}}
\toprule
\textbf{Capability} & \textbf{Function} & \textbf{Typical approaches} & SciForge differentiation \\
\midrule
Model Router
& Select the optimal LLM provider and model per request
& Single-provider lock-in or manual model switching
& Provider-agnostic routing across OpenAI, Anthropic, DeepSeek, and local models; cost-aware fallback chains; localizable deployment $\uparrow$ \\
Scientific Pipeline
& Translate scientific file formats into evidence the agent can reason over
& Raw text input or manual format conversion
& Four domain-expert translators (Esm2Text, Prot2Text, BioT5+, Cell2Sentence~(C2S)) covering protein, structure, molecule, and single-cell modalities; unrecognized formats fail closed with diagnostics $\uparrow$ \\
Agent Runtime
& Execute AI agents with tools, sub-agents, and workspace operations
& Single-session chat without structured delegation
& Multi-backend support (Codex, Claude Code, custom runtimes); sub-agent delegation; MCP worker invocation; long-task recovery $\uparrow$ \\
Workflow Engine
& Chain research steps into repeatable, scheduled protocols
& Ad-hoc scripts and cron jobs
& DAG-based workflows with code, agent, tool, and approval nodes; schedule and event triggers; integrated with agent runtime $\uparrow$ \\
Research Memory
& Store and retrieve project knowledge across sessions
& Ephemeral chat history or external document RAG
& Structured Project DAG with confidence-tagged, scoped free-text records; typed goals and decision events persist across sessions $\uparrow$ \\
Evidence Governance
& Track provenance, audit claims, and govern release decisions
& Outputs detached from data, scripts, and review history
& Thread-scoped Evidence DAGs with automatic provenance capture; deterministic audit runs; goal-scoped release gates $\uparrow$ \\
Research Capability Patterns
& Define structured human judgment points at research milestones
& Generic chat and file-list views
& Six recurrent interaction patterns (artifact review, approval, visual inspection, etc.) grounded in Evidence DAGs and audit records $\uparrow$ \\
\bottomrule
\end{tabular}
\end{table}
\clearpage

\section{Workbench Interface Examples}
\label{app:workbench-interfaces}

This appendix shows the running SciForge desktop interface rather than isolated architectural diagrams. The screenshots were captured from real local research sessions and illustrate how agent execution, evidence inspection, project-level continuity, and manuscript review are exposed as human judgment surfaces.

\subsection{Thread-Level Evidence DAG}

\begin{figure}[H]
\centering
\includegraphics[width=\textwidth]{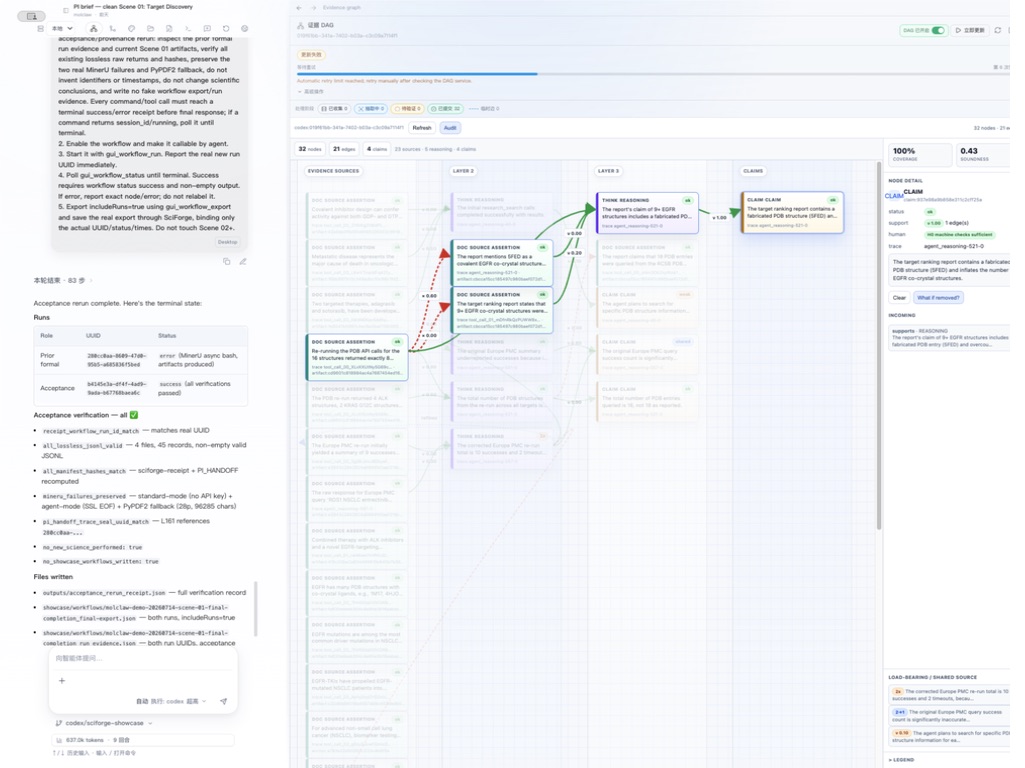}
\caption{Thread-level Evidence DAG embedded beside the originating SciForge research session. Source assertions, reasoning, claims, edge semantics, audit controls, and node-level provenance are visible in one review surface.}
\label{fig:appendix-evidence-dag}
\end{figure}

Figure~\ref{fig:appendix-evidence-dag} shows the Evidence DAG for a target-discovery session. The captured graph contains 32 nodes and 21 edges, including four explicit claims. It separates source assertions from reasoning and claim nodes, draws support and contradiction links between them, and exposes the selected node's provenance and incoming evidence in the inspector. In this example, the graph makes an overcounted and fabricated PDB reference directly inspectable instead of leaving the problem buried in a long agent transcript. The extraction and fragility summaries at the top provide a compact audit signal, while the underlying scientific judgment remains with the researcher.

\subsection{Project-Level DAG}

\begin{figure}[H]
\centering
\includegraphics[width=\textwidth]{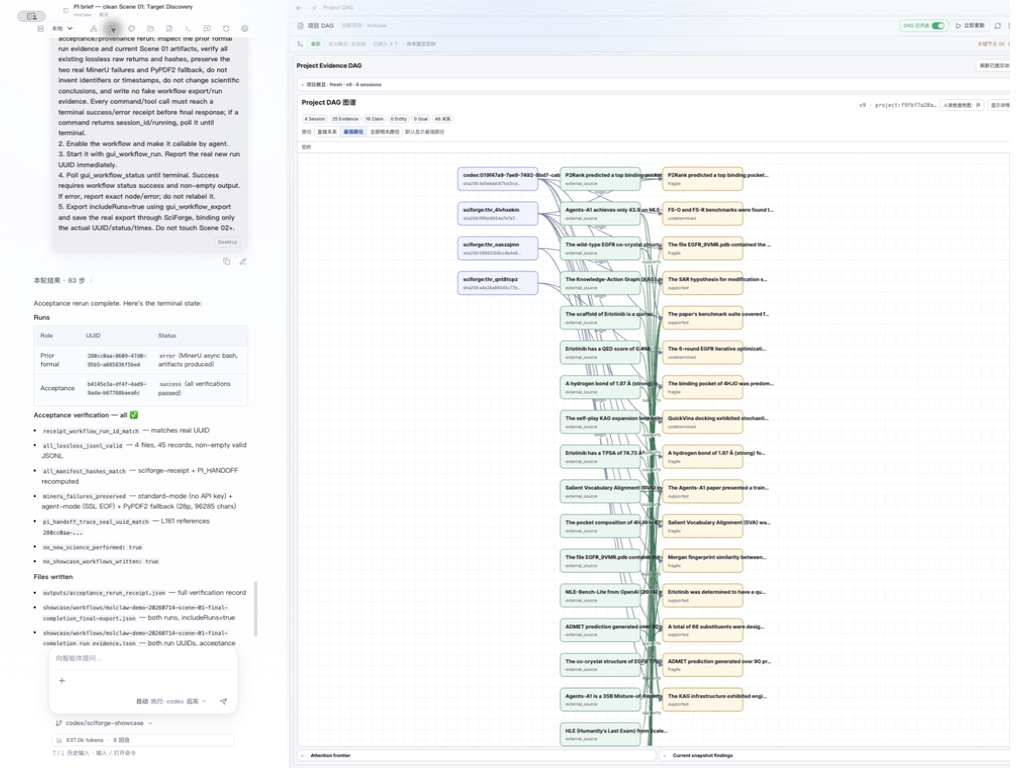}
\caption{Project DAG for the same research workspace. The committed project snapshot aggregates evidence and claims across sessions, exposing the project-level relationship structure and review diagnostics.}
\label{fig:appendix-project-dag}
\end{figure}

Figure~\ref{fig:appendix-project-dag} moves from a single thread to the project scope. The captured committed snapshot aggregates four sessions, 25 evidence records, and 16 claims into 45 nodes and 48 relationships. Session nodes identify the contributing runs, while evidence and claim nodes preserve how findings accumulate across the project. This view makes cross-session continuity reviewable and gives release and review workflows a shared project state rather than relying on chat history alone.

\clearpage
\subsection{Review-to-Revision PDF Workflow}

\begin{figure}[H]
\centering
\includegraphics[width=\textwidth]{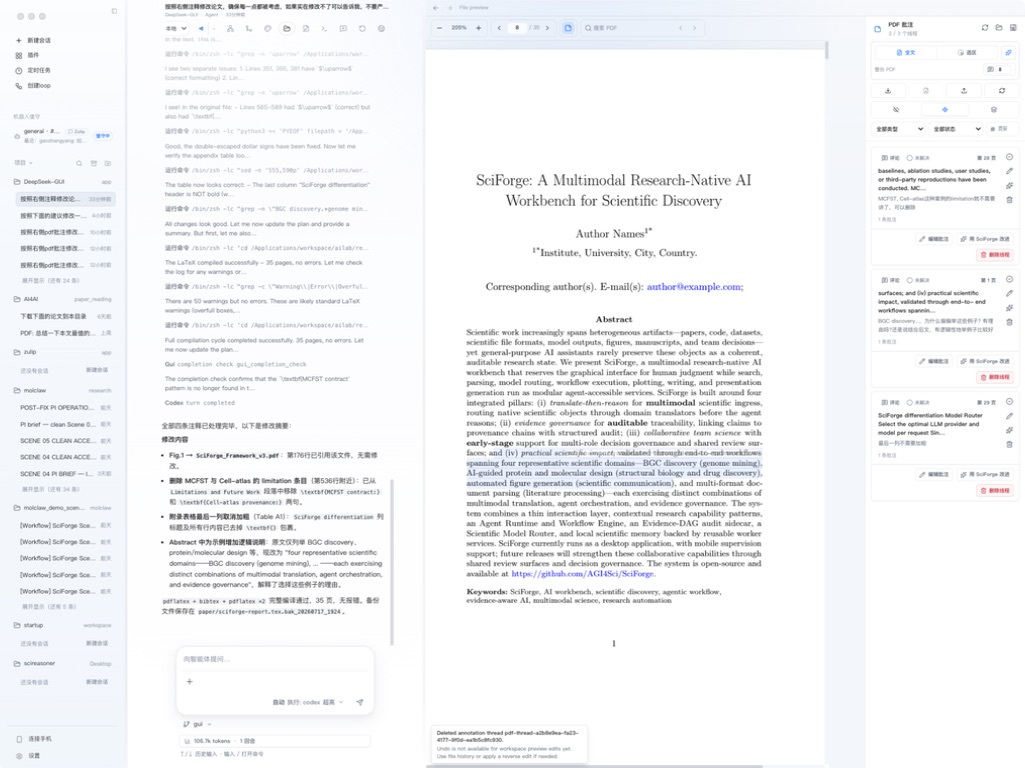}
\caption{Review-to-revision PDF workflow in SciForge. The agent thread records concrete manuscript changes and validation (left), the recompiled paper remains inspectable (center), and page-anchored reviewer comments retain their source context and improvement actions (right).}
\label{fig:appendix-pdf-review}
\end{figure}

SciForge also keeps human review close to the artifact being judged. Figure~\ref{fig:appendix-pdf-review} captures a completed manuscript-revision cycle rather than a static annotation list. Three page-anchored review threads question the abstract's example selection, the scope of the limitations discussion, and emphasis in an appendix table. The agent thread records the corresponding edits and compilation checks, while the recompiled manuscript and the still-open review items remain visible. This co-located view makes the path from a precise reviewer concern to an auditable artifact change immediately legible.

\clearpage
\subsection{Selection-Grounded Protein Structure Dialogue}

\begin{figure}[H]
\centering
\includegraphics[width=\textwidth]{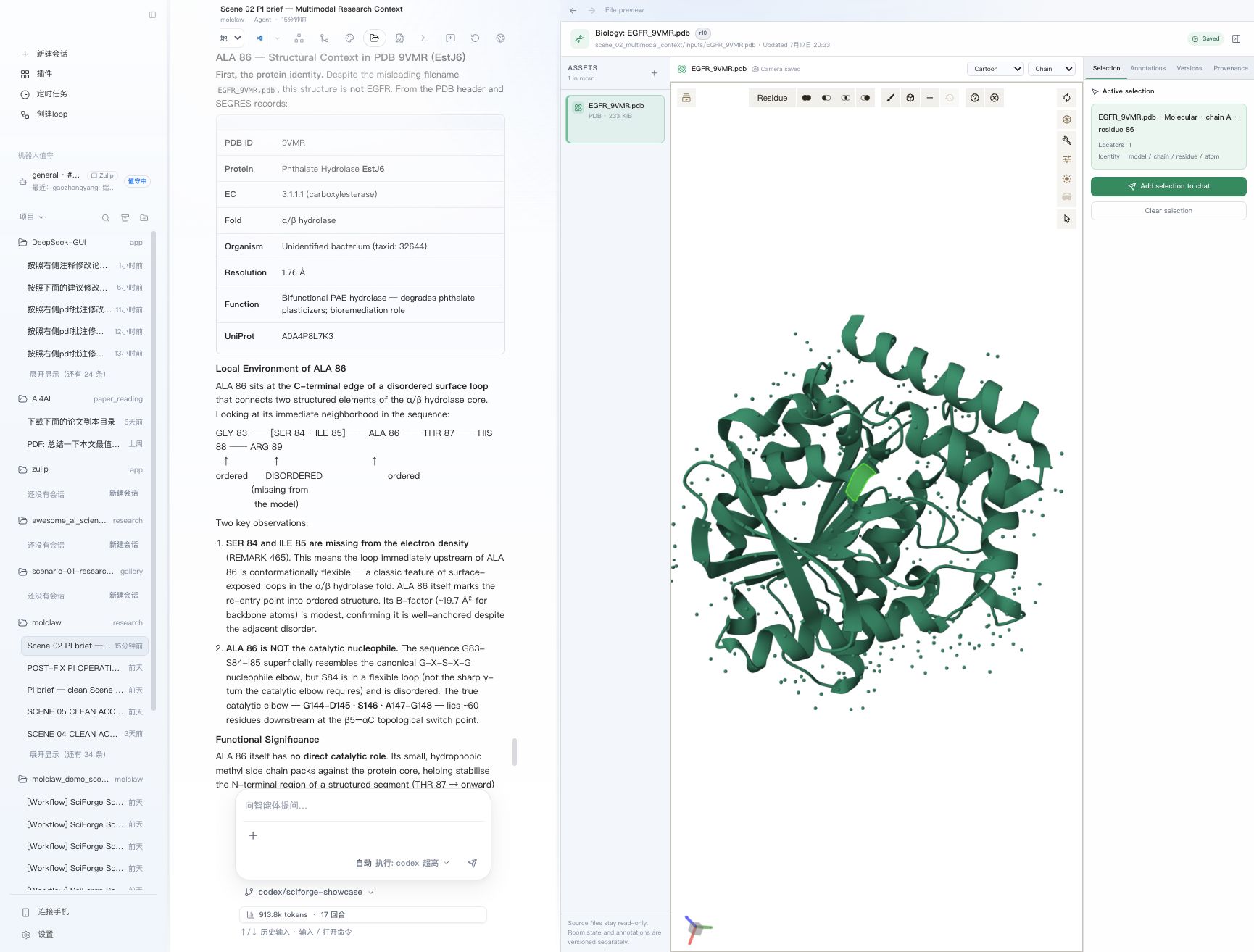}
\caption{Selection-grounded protein structure dialogue in SciForge. The Biology Room (right) retains the active ALA~86 selection in the 9VMR structure, while the linked chat (left) receives the file hash, room revision, and residue locator and uses them to drive an identity-checked structural interpretation.}
\label{fig:appendix-biology-selection-chat}
\end{figure}

Figure~\ref{fig:appendix-biology-selection-chat} shows a Biology Room selection connected directly to the chat composer. A residue picked in the three-dimensional viewer is represented as structured context containing the source-file path and SHA-256 digest, room revision, molecular identity fields, and an explicit model/chain/residue locator. Selecting \emph{Add selection to chat} appends this context to the current conversation, where the researcher can ask a follow-up question while keeping the structure, active selection, and resulting explanation visible together. In the captured session, the misleading filename is not accepted as biological identity: the follow-up explicitly requests identity verification before interpretation, and the agent checks the 9VMR record before discussing the selected ALA~86 environment and proposing additional residues for interactive inspection. This workflow turns a visual pick into an auditable, selection-grounded scientific dialogue rather than an unreferenced natural-language question.

\section{Author Contributions}

\begingroup
\small
\renewcommand{\arraystretch}{1.25}
\begin{tabular}{@{}L{0.23\textwidth} L{0.72\textwidth}@{}}
\textbf{Project Lead} & Zhangyang Gao\textsuperscript{1}. \\
\textbf{Core Contributors} & Zhangyang Gao\textsuperscript{1}; Minghao Fang\textsuperscript{2}; Yifei Liu\textsuperscript{2}; Hanhui Yang\textsuperscript{3}. \\
\textbf{Manuscript Preparation} & Xinyu Gu\textsuperscript{1}; Shixiang Tang\textsuperscript{1}; Siqi Sun\textsuperscript{1}; Lei Bai\textsuperscript{1}; Cheng Tan\textsuperscript{1}; Mengdi Liu\textsuperscript{4}; Hao Wu\textsuperscript{1}; Shuizhou Chen\textsuperscript{1}. \\
\end{tabular}

\medskip
\noindent\textbf{Use Case Responsibilities}
\begin{itemize}
  \item Agentic Research Sprint (Section~4.1) and AI-Guided Molecular Design (Section~4.7): Zhangyang Gao\textsuperscript{1}
  \item Reviewer / Rebuttal Mode (Section~4.3): Xinjie Mao\textsuperscript{1}
  \item Protein Structure Modeling (Section~4.2): Lang Yu\textsuperscript{1}
  \item Guided Paper Reproduction: MCFST (Section~4.4): Wenxuan Huang\textsuperscript{1}
  \item Cross-Scale Cell Atlas (Section~4.5): Xue Xia\textsuperscript{1}
  \item AI-Guided Protein Design (Section~4.6): Chengkai Yao\textsuperscript{1}
  \item Genome-to-BGC Discovery (Section~4.8): Wenjie Kang\textsuperscript{1}
\end{itemize}

\noindent\textbf{Affiliations}
\begin{itemize}
  \item[\textsuperscript{1}] Shanghai Artificial Intelligence Laboratory
  \item[\textsuperscript{2}] Zhejiang University
  \item[\textsuperscript{3}] Sichuan University
  \item[\textsuperscript{4}] Institute of Computing Technology, Chinese Academy of Sciences
\end{itemize}

\noindent\textbf{Correspondence:} Zhangyang Gao (\texttt{gaozhangyang@pjlab.org.cn}).
\endgroup

\end{appendices}
\clearpage

% ============================================================
% Resource
% ============================================================
\section*{Resource}

\noindent Use case implementations and associated artifacts are maintained in the following repositories:

\begin{itemize}
  \item \textbf{Agentic Research Sprint} (Section~4.1): \url{https://github.com/AGI4Sci/scenario-01-research-sprint} --- A multi-day, PI-controlled agentic loop for meiosis gene discovery with 132 stages and 199+ Git-tracked commits.
  \item \textbf{AI4AI: Protein and Structure Modeling} (Section~4.2): \url{https://github.com/BruthYU/autoresearch_base} --- Autonomous contact prediction via ESMC-6B ContactProbe with 24-iteration hyperparameter search and PDB-based evaluation.
  \item \textbf{Reviewer / Rebuttal Mode} (Section~4.3): \url{https://github.com/maoxinjie/scenario-05-reviewer-rebuttal-vcbench} --- Evidence-governed peer review and response workflow.
  \item \textbf{Guided Paper Reproduction: MCFST} (Section~4.4): \url{https://github.com/Winshion/sciforge-ai4ai-spacial-trans} --- End-to-end reproduction of MCFST spatial transcriptomics with generated code, preprocessed data, prediction outputs, and result figures.
  \item \textbf{Cross-Scale Cell Atlas} (Section~4.5): \url{https://github.com/ShaysXIA/cross-scale-data-demo} --- Multi-database integration and guided-level analysis for cross-scale cell atlas construction.
  \item \textbf{AI-Guided Protein Design} (Section~4.6): \url{https://github.com/kaiwinYao1/sciforge-de-novo-protein-demo} --- De novo protein design with ProteinMPNN sequence generation and Boltz-2 / ESMFold structural validation.
  \item \textbf{AI-Guided Molecular Design} (Section~4.7): \url{https://github.com/AGI4Sci/molclaw} --- Iterative molecular optimization over 4-anilinoquinazoline scaffold with scaffold-based SAR exploration, including 135 filtered candidates and 36 docking evaluations.
  \item \textbf{Genome-to-BGC Discovery} (Section~4.8): \url{https://github.com/wenne-kwj/scenario-bgc-genome-discovery} --- End-to-end antiSMASH/MIBiG/BiG-SCAPE pipeline with Candidate BGC Cards, multi-agent scientific analysis, and evidence-grounded prioritization across 430 BGC regions.
\end{itemize}

\noindent The SciForge platform itself is open source at \url{https://github.com/AGI4Sci/SciForge}.

\bibliography{sn-bibliography}

\end{document}